# On better training the infinite restricted Boltzmann machines

Xuan Peng, Xunzhang Gao, Xiang Li

College of Electronic Science, National University of Defense Technology

Changsha, China

E-Mails: pengxuan@nudt.edu.cn; gaoxunzhang@nudt.edu.cn; lixiang01@vip.sina.com

**Abstract** The infinite restricted Boltzmann machine (iRBM) is an extension of the classic RBM. It enjoys a good property of automatically deciding the size of the hidden layer according to specific training data. With sufficient training, the iRBM can achieve a competitive performance with that of the classic RBM. However, the convergence of learning the iRBM is slow, due to the fact that the iRBM is sensitive to the ordering of its hidden units, the learned filters change slowly from the left-most hidden unit to right. To break this dependency between neighboring hidden units and speed up the convergence of training, a novel training strategy is proposed. The key idea of the proposed training strategy is randomly regrouping the hidden units before each gradient descent step. Potentially, a mixing of infinite many iRBMs with different permutations of the hidden units can be achieved by this learning method, which has a similar effect of preventing the model from over-fitting as the dropout. The original iRBM is also modified to be capable of carrying out discriminative training. To evaluate the impact of our method on convergence speed of learning and the model's generalization ability, several experiments have been performed on the binarized MNIST and CalTech101 Silhouettes datasets. Experimental results indicate that the proposed training strategy can greatly accelerate learning and enhance generalization ability of iRBMs.

**Keywords** Infinite restricted Boltzmann machines; Model averaging; Regularization; Discriminative and generative training objective

## 1 Introduction

Boltzmann machines are stochastic neural networks consisting of symmetrically-coupled binary stochastic units [1]. They are proposed to find statistical regularities in the data. One of the most popular subset of Boltzmann machines is the restricted Boltzmann machine (RBM) [2]. The RBM and its various extensions have enjoyed much popularity for pattern analysis and generation. The generality and flexibility of RBMs enable them to be used in a wide range of applications, e.g., image and audio classification [3,4], generation [5], collaborative filtering [6,7], motion modeling [8] etc. More specifically, an RBM is a bipartite graphical model that uses a layer of hidden binary variables or units to model the probability distribution of a layer of visible variables. With enough hidden units, an RBM is able to represent a binary probability distribution fitting the training data as well as possible.

In general, adding more hidden units can improve the representational power of the model [9]. However, as the number of hidden units increases, many learned features become strongly correlated [10,11], which increases the risk of over fitting. Choosing a proper number of hidden units requires a procedure of model selection, which is time-consuming. To deal with this issue, Welling et al. [12] propose a boosting algorithm in the feature space of RBM, and at each iteration, a new feature is added and greedily learned. Nair，et al. [13] conceptually tie the weights of an infinite number of binary hidden units, and connect these sigmoid units with noisy rectified linear units (ReLUs) for better feature learning. More recently, Côté and Larochelle [14] have proposed a non-parametric model called the iRBM. By making the effective number of hidden units participating in the energy function change freely during training, the iRBM can automatically adjust the effective number of hidden units according to the data. The implicitly infinite capability of the iRBM also makes it capable of lifelong learning. Despite that, there is a major drawback for the iRBM, the slow convergence of learning, which offsets the advantage it brings to a certain degree. The reason for this drawback is that, the hidden units are correlated with each other given the visible variables. The learned filters or feature detectors change slowly from the left-most hidden unit to right, which is also called "ordering effect" in [14].

The fixed order of hidden units is the reason of strong dependency between filters learned by neighboring hidden units. The newly added hidden unit always begins to learns features jointly with the previous hidden units. In fact, the hidden units does not have to be constrained a fixed order, all possible permutations can be considered and evaluated. To achieve this, a random permutation of the hidden units is sampled from a certain distribution before each gradient descent step. Thus, the neighbors of each hidden unit are continuously changing as training progresses, which encourages the hidden units to learn features depending on themselves. By doing this, a different iRBM is trained at each gradient descent step. As there are infinite many hidden units in the iRBM, a mixture of infinite many iRBMs with different permutation of hidden units but shared parameters can be achieved theoretically. From this point of view, the proposed training strategy provides an effective way of preventing the model from over-fitting, as averaging of different models nearly always improves the generalization performance [15]. A similar effect can be achieved by dropout [16], which randomly drops units of the neural network during training, and is equivalent to combining exponentially many different sub networks, and also serves as regularization by an adaptive weight decay and sparse representation [17]. In fact, a more generalized iRBM is defined by treating the permutation of hidden units as a model parameter. Besides the new training strategy, another contribution of our work is extending the iRBM to being capable of performing supervised learning, which allows us to compare the generalization performance of our training strategy to other training methods and models on discrimination tasks.

**Related work** It needs to be mentioned here that, Ping, et al. [18] have proposed an alternative definition of the infinite RBM, and accordingly the Frank-Wolfe algorithm to learn it. We name their model FW-iRBM to avoid confusing with the model studied in this paper. The definition of FW-iRBM is motivated by the marginal log-likelihood of Classic RBMs. The weight matrix $\mathbf{W}$ is treated as $N$ samples of weight vector $\boldsymbol{\omega}$. The marginal log-likelihood of RBM is thus sampling approximation to marginal log-likelihood of the so-called "fractional RBM". Adding a hidden unit is equivalent to draw a new sample of $\boldsymbol{\omega}$ of fractional RBM. The training objective of fractional RBM tries to learning the distribution of $\boldsymbol{\omega}$, $q(\boldsymbol{\omega})$. They proposes a greedy training procedure which adds a hidden unit at each iteration and updates the weight of newly added hidden unit. The advantage of FW-iRBM is that it is a more generalized model that can be extended to an RBM with uncountable number of hidden units, as long as $q(\boldsymbol{\omega})$ is a continuous distribution. However, even though the order of hidden units invariant to model's marginal log-likelihood, the training procedure of FW-iRBM indicates that the order of hidden units still has an effect on the final performance. The reason is that at each step only the parameter of newly added hidden unit is updated based on all the previous added hidden units. And this greedy optimization algorithm is more likely to result in a sub-optimal solution, that is the reason that the performance is not monotonically improving as the model size gets larger. The iRBM doesn't encounter this problem as it simultaneously updates all the non-zero parameters and automatically decides whether adding a new hidden unit or not.

The remainder of this paper is organized in the following order: Sect.2 gives a brief review of the iRBM model, after which the discriminative iRBM is introduced, and the cause of ordering effect is briefly analyzed. In Sect. 3, the proposed training strategy is formally presented, and an condition under which the model is invariant to the order of hidden units is proposed with a proof in the appendix. And in Sect. 4, several experiments are performed to empirically evaluate our training strategy. Finally, we conclude our work in Sect. 5.

## 2 Infinite restricted Boltzmann machines

In this section, the original iRBM is first introduced briefly. After that, some modifications to the energy function of the iRBM are made, which leads to the discriminative iRBMs. Finally, we briefly analyze cause of the ordering effect in iRBMs.

## 2.1 iRBM

The iRBM [14] is proposed to address the difficulty of deciding a proper size of the hidden layer for the RBM, it can effectively adapt its capacity as training progresses.

The iRBM is an extension of the classic RBM, which mixes infinitely many RBMs with different number of hidden units from 1 to $\infty$, and all the RBMs choose the hidden units from the same sequence starting from the first one. It defines a probability $p(\mathbf{v},\mathbf{h},z)$ with the energy function given as follows:

$$E(\mathbf{v},\mathbf{h},z) = -\mathbf{v}^T\mathbf{b}^v - \sum_{i=1}^{z} h_i\left(\mathbf{W}_{i\cdot}\mathbf{v} + b_i^h\right) - \beta_i , \tag{1}$$

where, $\mathbf{v} \in \{0,1\}^D$ is the $D$ dimensional visible vector representing the observable data. $h_i \in \{0,1\}$ is the $i^{\text{th}}$ element of the infinite-dimensional hidden vector $\mathbf{h}$. The random variable $z \in \mathrm{N}$ can be understood as the total number of hidden units being selected to participate in the energy function. $\beta_i$ is the penalty for each selected hidden unit $h_i$. $\mathbf{W}_{i\cdot}$ is the $i^{\text{th}}$ row of the weight matrix $\mathbf{W}$ connecting the visible units and the hidden units. $\mathbf{b}^v$ is the visible units bias vector. $b_i^h$ is the $i^{\text{th}}$ hidden unit bias.

The joint probability over $\mathbf{v}$, $\mathbf{h}$ and $z$ is

$$p(\mathbf{v},\mathbf{h},z) = \frac{1}{Z} e^{-E(\mathbf{v},\mathbf{h},z)} , \tag{2}$$

where

$$Z = \sum_{z'=1}^{\infty} \sum_{\mathbf{v}'} \sum_{\mathbf{h}' \in H_{z'}} e^{-E(\mathbf{v}',\mathbf{h}',z')} , \tag{3}$$

and $H_z = \{h \in H \mid h_k = 0 \forall k > z\}$ , where $H$ is the set of all possible values $\mathbf{h}$ takes. Thus $H_z$ defines the legal values of $\mathbf{h}$ given $z$. The graphical model of the iRBM is shown in Fig. 1.

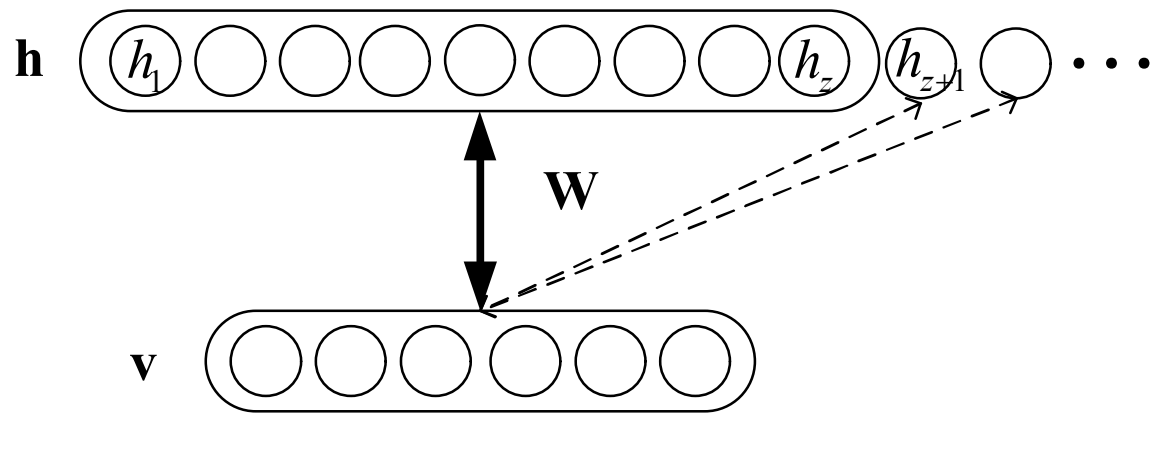

Fig.1. Graphical model of the iRBM.

It should be noticed that, for a given $z$, the value of the energy function is irrelevant for the dimensions of $\mathbf{h}$ from $z+1$ to $\infty$, which means that $h_i$ where $i > z$ will never be activated. Thus, (1) has the same form with the energy function of the classic RBM with $z$ hidden units except the penalty $\beta_i$ which the latter does not have.

By marginalizing out $\mathbf{h}$ and $z$, the marginal distribution $p(\mathbf{v})$ is derived as follows:

$$p(\mathbf{v}) = \frac{1}{Z}\sum_{z=1}^{\infty}\sum_{\mathbf{h}\in H_z} e^{-E(\mathbf{v},\mathbf{h},z)} = \frac{1}{Z}\sum_{z=1}^{\infty} e^{-F(\mathbf{v},z)} , \tag{4}$$

where

$$F(\mathbf{v},z) = -\mathbf{v}^T\mathbf{b}^v - \sum_{i=1}^{z}\ln\left(1+\exp\left(\mathbf{W}_{i\cdot}\mathbf{v}+b_i^h\right)\right) - \beta_i . \tag{5}$$

The iRBM is trained by maximizing the likelihood of the training examples, the objective function is defined below:

$$f(\Theta, D_{\text{train}}) = \frac{1}{|D_{\text{train}}|}\sum_{n=1}^{|D_{\text{train}}|} -\ln p(\mathbf{v}_n) , \tag{6}$$

where $D_{\text{train}}$ is the set of training examples.
The gradient of this objective function has a simple form given below:

$$\frac{\partial}{\partial\theta} f(\Theta, D_{\text{train}}) = \frac{1}{|D_{\text{train}}|}\sum_{n=1}^{|D_{\text{train}}|}\left(\mathbb{E}_{p(z|\mathbf{v}_n)}\left[\frac{\partial}{\partial\theta}F(\mathbf{v}_n,z)\right] - \mathbb{E}_{p(z,\mathbf{v})}\left[\frac{\partial}{\partial\theta}F(\mathbf{v},z)\right]\right) , \tag{7}$$

where $\mathbb{E}_{p(x)}\left[f(x)\right]$ represents the expectation of $f(x)$ over distribution $p(x)$.

For nearly all types of BMs, computing the partition function $Z$ is intractable, therefore, the gradients of $f(\Theta, D_{\text{train}})$ cannot be exactly computed. This is also the case for the iRBMs, as there are infinite many hidden units. Côté and Larochelle [14] suggest using the Contrastive Divergence (CD) and the Persistent Contrastive Divergence (PCD) [19,20] algorithms to approximately calculate the gradients, i.e. a Gibbs sampling is used to sample $(z,\mathbf{v}) \sim p(z,\mathbf{v})$, and then the second expectation in (11) is estimated with these samples. The $k^{\text{th}}$ Gibbs step is done in the following order: $z^{(k)} \sim p\left(z \mid \mathbf{v}^{(k-1)}\right) \to \mathbf{h}^{(k)} \sim p\left(\mathbf{h} \mid \mathbf{v}^{(k-1)}, z^{(k)}\right) \to \mathbf{v}^{(k)} \sim p\left(\mathbf{v} \mid \mathbf{h}^{(k)}, z^{(k)}\right)$.The detail of learning is provided in [14].

The hidden units of iRBM are selected in sequence as $z$ takes the value from 1 to $\infty$, and if the penalty $\beta_i$ is chosen properly, an infinite pool of hidden units can be achieved. A way to parameterize $\beta_i$ is suggested in [14], which is $\beta_i = \beta\ln\left(1+e^{b_i^h}\right)$. This will ensure the infinite summing in partition function $Z$ is convergent as long as $\beta > 1$ and the number of hidden units having non-zero weights and biases is always finite.

## 2.2 Discriminative iRBM

To apply iRBM to discrimination tasks, it is possible to modify the energy function to make it target dependent just as the case of Classification RBM [21]. And we call the new model discriminative infinite restricted Boltzmann machine (Dis-iRBM). The energy function of the Dis-iRBM is defined below:

$$E(\mathbf{v},\mathbf{h},y,z) = -\mathbf{v}^T\mathbf{b} - \mathbf{e}_y{}^T\mathbf{d} - \sum_{i=1}^{z} h_i\left(\mathbf{W}_{i\cdot}\mathbf{v} + \mathbf{U}_{i\cdot}\mathbf{e}_y + c_i\right) - \beta_i \quad , \tag{8}$$

where, $\mathbf{e}_y = \left(1_{i=y}\right)_{i=1}^{C}$ is the "one hot" representation of the label $y \in \{1,2,\cdots,C\}$. $\mathbf{d}$ is the label bias vector. $\mathbf{U}_{i\cdot}$ is the $i^{\text{th}}$ row of the weight matrix $\mathbf{U}$ connecting $\mathbf{h}$ and $\mathbf{e}_y$.

With the energy function defined above, the joint probability over $\mathbf{v}$, $\mathbf{h}$, $y$ and $z$ is given below:

$$p(\mathbf{v},\mathbf{h},y,z) = \frac{1}{Z} e^{-E(\mathbf{v},\mathbf{h},y,z)}, \tag{9}$$

where $Z = \sum_{z'=1}^{\infty}\sum_{y'}\sum_{\mathbf{v}'}\sum_{\mathbf{h}'\in H_{z'}} e^{-E(\mathbf{v}',\mathbf{h}',z',y')}$.

By marginalizing out $\mathbf{h}$ and $z$, we get the marginal distribution $p(\mathbf{v},y)$ as follows:

$$p(\mathbf{v},y) = \frac{1}{Z}\sum_{z=1}^{\infty}\sum_{\mathbf{h}\in H_z} e^{-E(\mathbf{v},\mathbf{h},y,z)} = \frac{1}{Z}\sum_{z=1}^{\infty} e^{-F(\mathbf{v},y,z)}, \tag{10}$$

where

$$F(\mathbf{v},y,z) = -\mathbf{v}^T\mathbf{b} - \mathbf{e}_y{}^T\mathbf{d} - \sum_{i=1}^{z} \ln\left(1 + \exp\left(\mathbf{W}_{i\cdot}\mathbf{v} + \mathbf{U}_{i\cdot}\mathbf{e}_y + c_i\right)\right) - \beta_i \quad . \tag{11}$$

The following conditional distributions can be easily derived from the energy function (8):

$$p(h_i = 1 \mid \mathbf{v},z,y) = \begin{cases} \mathrm{s}\left(\mathbf{w}_{i\cdot}\mathbf{v} + \mathbf{U}_{i\cdot}\mathbf{e}_y + c_i\right) & , i \le z \\ 0 & , i > z \end{cases}, \tag{12}$$

$$p(v_j = 1 \mid \mathbf{h},z) = \mathrm{s}\left(b_i + \sum_{i=1}^{z} h_i W_{ij}\right) \quad , \tag{13}$$

$$p(y \mid \mathbf{h},z) = \exp\left(d_y + \sum_{i=1}^{z} h_i\left(\mathbf{U}_{i\cdot}\mathbf{e}_y\right)\right) \Big/ \sum_{y'}\exp\left(d_{y'} + \sum_{i=1}^{z} h_i\left(\mathbf{U}_{i\cdot}\mathbf{e}_{y'}\right)\right) \quad , \tag{14}$$

where $\mathrm{s}(x) = 1/\left(1+e^{-x}\right)$ is the sigmoid function.

The graphical model of Dis-iRBM is shown in Fig. 2.

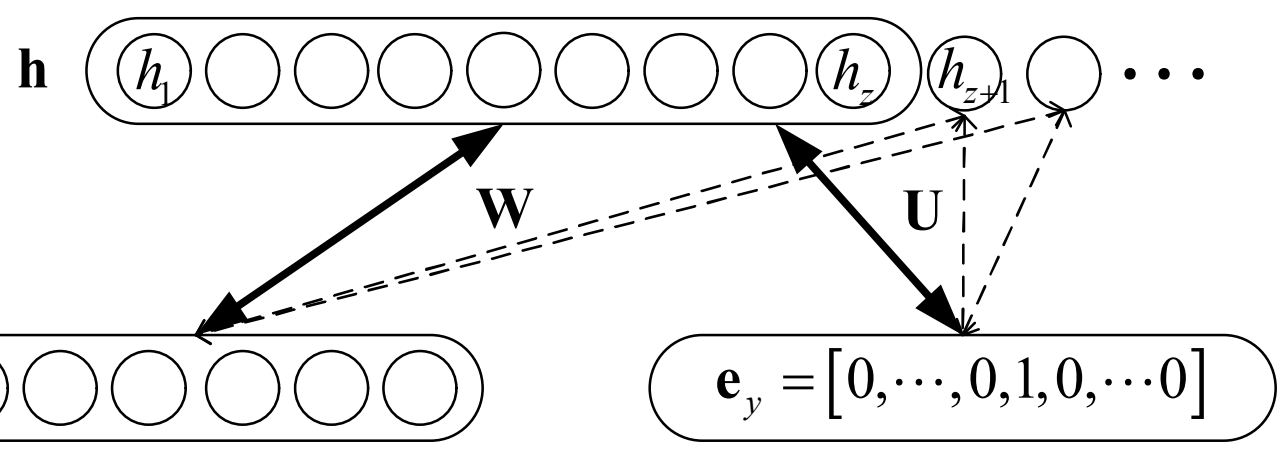

Fig. 2. Graphical model of the Dis-iRBM.

## 2.3 The ordering effect in iRBMs

The reason for slow convergence of learning for iRBMs is that newly added hidden units are correlated to previous hidden units, it takes a long time for the filters to become diverse from each other.

To briefly explain this phenomenon, let's firstly take a look at the conditional probability of the $i^{\text{th}}$ hidden unit:

$$\begin{aligned} p(h_i = 1 \mid \mathbf{v}) &= \sum_{z=i}^{\infty} p(h_i = 1 \mid \mathbf{v}, z) p(z \mid \mathbf{v}) = \sum_{z=i}^{\infty} \mathrm{s}(\mathbf{W}_{i\cdot}\mathbf{v} + c_i) p(z \mid \mathbf{v}) \\ &= \mathrm{s}(\mathbf{W}_{i\cdot}\mathbf{v} + c_i) p(z \ge i \mid \mathbf{v}) \end{aligned} , \quad (15)$$

and

$$p(z \ge i \mid \mathbf{v}) = \sum_{z=i}^{\infty} p(z \mid \mathbf{v}) = \sum_{z=i}^{\infty} \frac{\exp(-F(\mathbf{v}, z))}{\sum_{z'=1}^{\infty} \exp(-F(\mathbf{v}, z'))} . \quad (16)$$

It can be seen from (16) that computing $p(z \ge i \mid \mathbf{v})$ involves summing up all possible states of the other hidden units $\mathbf{h}$. The conditional probability $p(h_i = 1 \mid \mathbf{v})$ is a multiply of two terms, one of which is the sigmoid function identical to that of the classic RBM, the other reflects the influence of the other hidden units. The hidden units are co-adapted to represent features. This co-adaption sometimes is harmful to generalization.

Suppose $i = l_t$ which corresponds to the newly added hidden unit at step $t$. Now,

$$\begin{aligned} p(h_{lt} = 1 \mid \mathbf{v}) &= \mathrm{s}(\mathbf{W}_{lt\cdot}\mathbf{v} + c_i) p(z \ge l_t \mid \mathbf{v}) \\ &= \mathrm{s}(\mathbf{W}_{lt\cdot}\mathbf{v} + c_i)\left(p(z = l_t \mid \mathbf{v}) + \text{constant}\right) \end{aligned} . \quad (17)$$

The constant comes from the infinite sum of all the remaining hidden units with zero parameters.

Equation (17) indicates that, the newly added hidden unit is always influenced by all the previous added hidden units. It starts to learn the feature jointly with the previous features not independently by itself.

## 3 The proposed training strategy

In this section, the proposed training strategy is formally presented, which consists of two parts, the dynamic training objective and the approximated gradient descent algorithm for optimizing the objective.

### 3.1 The dynamic training objective

The basic training objective studied in this paper is a hybrid training objective combining the generative and the discriminative training objective similar to [21], and is given below:

$$\begin{aligned} f_{\text{hybrid}}\left(\Theta, D_{\text{train}}\right) &= -\frac{1}{\left|D_{\text{train}}\right|}\left(\left(1+\alpha\right)\sum_{n=1}^{\left|D_{\text{train}}\right|} \ln p\left(y_n \mid \mathbf{v}_n\right) + \alpha \sum_{n=1}^{\left|D_{\text{train}}\right|} \ln p\left(\mathbf{v}_n\right)\right) \\ &= \frac{1}{\left|D_{\text{train}}\right|}\left(\left(1+\alpha\right) f_{\text{dis}}\left(\Theta, D_{\text{train}}\right) + \alpha f_{\text{gen}}\left(\Theta, D_{\text{train}}\right)\right) \end{aligned}, \quad (18)$$

where, $D_{\text{train}} = \left\{\left(\mathbf{v}_n, y_n\right)\right\}$ is the set of training examples. $f_{\text{dis}}\left(\Theta, D_{\text{train}}\right) = -\sum_{n=1}^{\left|D_{\text{train}}\right|} \ln p\left(y_n \mid \mathbf{v}_n\right)$ corresponds to the discriminative part modeling $p\left(y_n \mid \mathbf{v}_n\right)$, and $f_{\text{gen}}\left(\Theta, D_{\text{train}}\right) = -\sum_{n=1}^{\left|D_{\text{train}}\right|} \ln p\left(\mathbf{v}_n\right)$ corresponds to the generative part modeling the inputs $p\left(\mathbf{v}_n\right)$ only. $\alpha > 0$ controls the proportion of each part. The second part can be thought of as a model regularization term.

Stochastic gradient descent or mini-batch gradient descent method can be used to minimize the objective function (18). However, as mentioned above, convergence of learning is slow for iRBMs.

An interesting fact can be derived from (15) that, if $p\left(z < i \mid \mathbf{v}\right) \to 0$, equation (15) becomes

$$p\left(h_i = 1 \mid \mathbf{v}\right) = \mathrm{s}\left(\mathbf{W}_{i\cdot}\mathbf{v} + c_i\right) . \quad (19)$$

In this case, the $i^{\text{th}}$ hidden unit is independent with the other hidden units. This indicates that, if $p\left(z < M \mid \mathbf{v}\right) \to 0$, then the first $M$ hidden units are independent with each other and act like a classic RBM, and any order of the first $M$ hidden units does not influence the performance of the model.

In [14] the iRBM is trained in a fixed order, which leads to an order biased model.

If we assume the order of hidden units is changeable, and jointly train iRBMs with all possible orders, the bias might be alleviated, and the learned features will become closer to (19). This inspires us to propose an alternative training objective as follows.

At each gradient descent step $t$, we first draw a sample of permutation $\tilde{\mathbf{o}}_t$ of the left-most $M_t$ hidden units from a distribution $p_t(\mathbf{o}_t)$:

$$\tilde{\mathbf{o}}_t = [\tilde{o}_1, \tilde{o}_2, \cdots, \tilde{o}_{Mt}] \sim p_t(\mathbf{o}_t),$$

where, $p_t(\mathbf{o}_t)$ is a discrete distribution as there are in total $M_t!$ permutations of $M_t$ numbers. In this paper, we use a uniform distribution $p_t(\mathbf{o}_t) = 1/(M_t!)$, which gives equal chance to each permutation. Other distributions can also be used. We note $l_t$ as the maximal number of activated hidden units at gradient descent step $t$. To stabilize the growing of selected hidden units, only part of them are regrouped ($M_t < l_t$).

Then, the left-most $M_t$ hidden units are reordered according to $\tilde{\mathbf{o}}_t$:

$$[h_1^t, h_2^t, \cdots, h_{Mt}^t] \leftarrow [h_{\tilde{o}_1}, h_{\tilde{o}_2}, \cdots, h_{\tilde{o}_{Mt}}],$$

and the parameters with each hidden unit are also reordered:

$$[\theta_1^t, \cdots, \theta_{Mt}^t] \leftarrow [\theta_{\tilde{o}_1}^t, \cdots, \theta_{\tilde{o}_{Mt}}^t].$$

Now, all probabilities are conditioned on permutation $\tilde{\mathbf{o}}_t$: $p(y_n \mid \mathbf{v}_n; \tilde{\mathbf{o}}_t)$ and $p(\mathbf{v}_n \mid \tilde{\mathbf{o}}_t)$.

Finally, the gradient of objective function (18) with $p(y_n \mid \mathbf{v}_n)$ and $p(\mathbf{v}_n)$ replaced by $p(y_n \mid \mathbf{v}_n; \tilde{\mathbf{o}}_t)$ and $p(\mathbf{v}_n \mid \tilde{\mathbf{o}}_t)$ is computed:

$$\frac{\partial}{\partial \theta^t} f_{\text{hybrid}}(\Theta_t, D_{\text{train}}, t) = \frac{1}{|D_{\text{train}}|} \frac{\partial}{\partial \theta^t} \left( (1+\alpha) f_{\text{dis}}(\Theta_t, D_{\text{train}}, t) + \alpha f_{\text{gen}}(\Theta_t, D_{\text{train}}, t) \right), \quad (20)$$

where,

$$f_{\text{dis}}(\Theta_t, D_{\text{train}}, t) = -\sum_{n=1}^{|D_{\text{train}}|} \ln p(y_n \mid \mathbf{v}_n; \tilde{\mathbf{o}}_t),$$

and

$$f_{\text{gen}}(\Theta_t, D_{\text{train}}, t) = -\sum_{n=1}^{|D_{\text{train}}|} \ln p(\mathbf{v}_n \mid \tilde{\mathbf{o}}_t).$$

$f_{\text{hybrid}}(\Theta_t, D_{\text{train}}, t)$ is the objective function at step $t$, its gradient (20) is in fact a

sampling approximation to the gradient of the following marginalized objective function:

$$F_{\text{hybrid}}\left(\Theta_t, D_{\text{train}}, t\right) = \frac{1}{\left|D_{\text{train}}\right|}\left(\left(1+\alpha\right) F_{\text{dis}}\left(\Theta_t, D_{\text{train}}, t\right) + \alpha F_{\text{gen}}\left(\Theta_t, D_{\text{train}}, t\right)\right) \quad , \quad (21)$$

where,

$$F_{\text{dis}}\left(\Theta_t, D_{\text{train}}, t\right) = -\sum_{n=1}^{\left|D_{\text{train}}\right|} \ln \sum_{\mathbf{o}_t} p\left(y_n \mid \mathbf{v}_n; \mathbf{o}_t\right) p_t\left(\mathbf{o}_t\right),$$

and

$$F_{\text{gen}}\left(\Theta_t, D_{\text{train}}, t\right) = -\sum_{n=1}^{\left|D_{\text{train}}\right|} \ln \sum_{\mathbf{o}_t} p\left(\mathbf{v}_n \mid \mathbf{o}_t\right) p_t\left(\mathbf{o}_t\right).$$

The word "dynamic" is used to emphasize that, the training objective changes accordingly as the number of regrouping hidden units changes. The proposed training procedure is briefly illustrated Fig. 3.

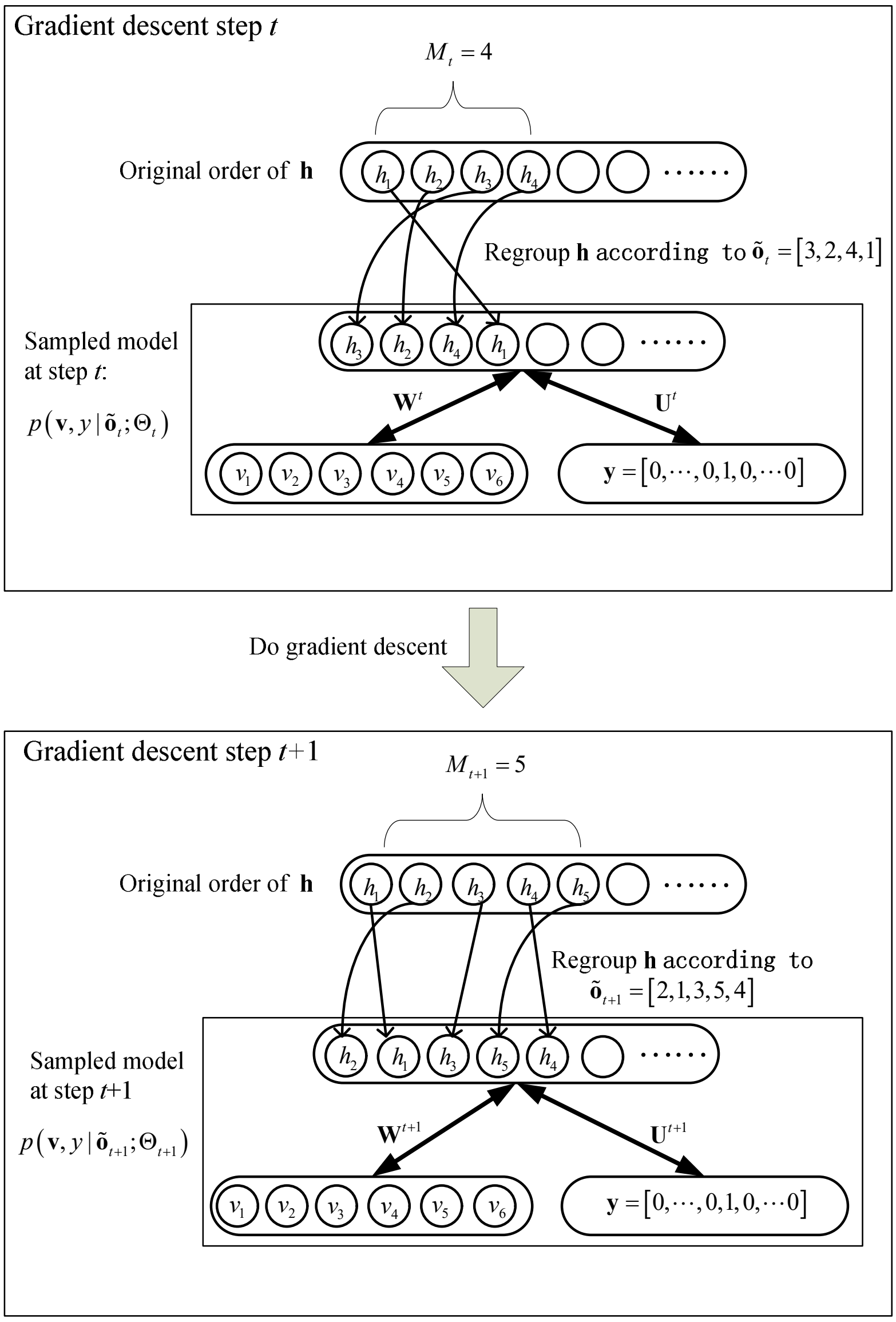

Fig. 3 An illustration of RP training at gradient descent step $t$ and $t$+1.

From the dynamic objective function (21), we can see that a mixture of $M_t!$ iRBMs is been trained at gradient descent step $t$. All the iRBMs share the same set of hidden units but with different permutations. When the number of hidden units with non-zero weights $l_t$ grows during training, the number of mixed iRBMs grows accordingly. Theoretically, $l_t$ is allowed to take arbitrary positive integer value. Thus, the proposed training objective can potentially allow a mixture of infinite many iRBMs. For convenience, we name our training strategy as "random permutation (RP)".

$M_t$ controls the proportion of hidden units being regrouped. If $M_t$ is too large, the model will grow explosively. If $M_t$ is too small, the boosting effect is minor. We prefer as many hidden units to be regrouped as possible, in the meanwhile, the model would not growing too rapidly. The strategy of choosing a proper $M_t$ will be specifically discussed in Sec. 4.

### 3.2 The approximated gradient descent algorithm

In the previous subsection we have defined the objective function (21) together with the approximated gradient (20) for parameter learning. In this subsection, we will provide a way of computing the gradient (20).

At each step $t$, the dynamic gradient (20) is identical to the gradient of the objective function (18) except the probability $p(y_n \mid \mathbf{v}_n)$ and $p(\mathbf{v}_n)$ are replaced by $p(y_n \mid \mathbf{v}_n; \tilde{\mathbf{o}}_t)$ and $p(\mathbf{v}_n \mid \tilde{\mathbf{o}}_t)$. Once $\tilde{\mathbf{o}}_t$ is sampled, the training objective aims at learning an iRBM with permutation $\tilde{\mathbf{o}}_t$ of the hidden units. The learning of the generative part $f_{\text{gen}}(\Theta_t, D_{\text{train}}, t)$ is identical to the learning of the iRBM, which is briefly introduced in Sect.2. CD or PCD can be directly used to compute the gradients. The approximated gradient for the generative part is given below:

$$\frac{\partial}{\partial \theta^t} f_{\text{gen}}(\Theta_t, D_{\text{train}}, t) \approx \frac{1}{|D_{\text{train}}|} \sum_{n=1}^{|D_{\text{train}}|} \left( \frac{\partial}{\partial \theta} F(\mathbf{v}_n, z^{\text{pos}} | \tilde{\mathbf{o}}_t) - \frac{\partial}{\partial \theta} F(\mathbf{v}^{\text{neg}}, z^{\text{neg}} | \tilde{\mathbf{o}}_t) \right) , \quad (22)$$

where, $z^{\text{pos}}$ is sampled from $p(z \mid \mathbf{v}_n, \tilde{\mathbf{o}}_t)$, and $\mathbf{v}^{\text{neg}}, z^{\text{neg}}$ are sampled from $p(\mathbf{v}, z | \tilde{\mathbf{o}}_t)$ by K-step Gibbs sampling.

The pseudo code of model parameter update for generative training objective is summarized in Algorithm 1，which is shown in the **Appendix**.

In order to calculate the gradient of the discriminative part $f_{\text{dis}}(\Theta_t, D_{\text{train}}, t)$, the conditional probability $p(y \mid \mathbf{v}; \tilde{\mathbf{o}}_t)$ is derived as follows:

$$p(y \mid \mathbf{v}, \tilde{\mathbf{o}}_t) = \frac{p(y, \mathbf{v} \mid \tilde{\mathbf{o}}_t)}{p(\mathbf{v} \mid \tilde{\mathbf{o}}_t)} = \frac{\sum_{z=1}^{\infty} e^{-F(\mathbf{v}, y, z \mid \tilde{\mathbf{o}}_t)}}{\sum_{z=1}^{\infty} \sum_{y'} e^{-F(\mathbf{v}, y', z \mid \tilde{\mathbf{o}}_t)}} = \frac{\sum_{z=1}^{\infty} e^{-G(y, z \mid \mathbf{v}; \tilde{\mathbf{o}}_t)}}{\sum_{z=1}^{\infty} \sum_{y'} e^{-G(y', z \mid \mathbf{v}; \tilde{\mathbf{o}}_t)}}, \quad (23)$$

where

$$G(y, z \mid \mathbf{v}; \tilde{\mathbf{o}}_t) = -d_y - \sum_{i=1}^{z} \ln\left(1 + \exp\left(\mathbf{W}_{i\cdot}^t \mathbf{v} + \mathbf{U}_{i\cdot}^t \mathbf{e}_y + c_i\right)\right) - \beta_i . \quad (24)$$

By taking equations (23) and (24) into the discriminative part of (20), the gradient of $f_{\text{dis}}(\Theta_t, D_{\text{train}}, t)$ is derived as follows:

$$\frac{\partial}{\partial \theta^t} f_{\text{dis}}(\Theta_t, D_{\text{train}}, t) \quad (25)$$

$$= \frac{1}{|D_{\text{train}}|} \sum_{n=1}^{|D_{\text{train}}|} \left( \mathbb{E}_{p(z \mid \mathbf{v}_n, y_n; \tilde{\mathbf{o}}_t)} \left[ \frac{\partial}{\partial \theta} G(y_n, z \mid \mathbf{v}_n; \tilde{\mathbf{o}}_t) \right] - \mathbb{E}_{p(z, y \mid \mathbf{v}_n; \tilde{\mathbf{o}}_t)} \left[ \frac{\partial}{\partial \theta} G(y, z \mid \mathbf{v}_n; \tilde{\mathbf{o}}_t) \right] \right),$$

and

$$p(z \mid \mathbf{v}, y; \tilde{\mathbf{o}}_t) = e^{-G(y, z \mid \mathbf{v}; \tilde{\mathbf{o}}_t)} \Big/ \sum_{z'=1}^{\infty} e^{-G(y, z' \mid \mathbf{v}; \tilde{\mathbf{o}}_t)}, \quad (26)$$

$$p(z, y \mid \mathbf{v}; \tilde{\mathbf{o}}_t) = e^{-G(y, z \mid \mathbf{v}; \tilde{\mathbf{o}}_t)} \Big/ \sum_{z'=1}^{\infty} \sum_{y'} e^{-G(y', z' \mid \mathbf{v}; \tilde{\mathbf{o}}_t)} . \quad (27)$$

For the same reason stated for training the iRBM, the (P)CD algorithm can also be used to approximate (25). Sampling $z, y \sim p(z, y \mid \mathbf{v}; \tilde{\mathbf{o}}_t)$ is done by repeating the two steps: $z^{(k)} \sim p(z \mid y^{(k-1)}, \mathbf{v}; \tilde{\mathbf{o}}_t)$ and then $y^{(k)} \sim p(y \mid z^{(k)}, \mathbf{v}; \tilde{\mathbf{o}}_t)$. $p(z \mid \mathbf{v}, y; \tilde{\mathbf{o}}_t)$ is already given in (26) and $p(y \mid z, \mathbf{v}; \tilde{\mathbf{o}}_t)$ is given below:

$$p(y \mid \mathbf{v}, z; \tilde{\mathbf{o}}_t) = e^{-G(y, z \mid \mathbf{v}; \tilde{\mathbf{o}}_t)} \Big/ \sum_{y'} e^{-G(y', z \mid \mathbf{v}; \tilde{\mathbf{o}}_t)} . \quad (28)$$

By taking sampling approximations, the gradient of the discriminative part can be replaced by the following equation:

$$\frac{\partial}{\partial \theta^t} f_{\text{dis}}(\Theta_t, D_{\text{train}}, t) \approx \frac{1}{|D_{\text{train}}|} \sum_{n=1}^{|D_{\text{train}}|} \left( \frac{\partial}{\partial \theta} G(y_n, z^{\text{pos}} \mid \mathbf{v}_n; \tilde{\mathbf{o}}_t) - \frac{\partial}{\partial \theta} G(y^{\text{neg}}, z^{\text{neg}} \mid \mathbf{v}_n; \tilde{\mathbf{o}}_t) \right), \quad (29)$$

Where $z^{\text{pos}}$ is a sample from $p(z \mid \mathbf{v}_n, y_n, \tilde{\mathbf{o}}_t)$, $y^{\text{neg}}$ and $z^{\text{neg}}$ are samples from $p(z, y \mid \mathbf{v}_n, \tilde{\mathbf{o}}_t)$ by K-step Gibbs sampling.

However, the approximated gradient (29) causes high variance during learning in practice. To alleviate this problem, an alternative derivation of the discriminative part is proposed as follows:

$$\frac{\partial}{\partial\theta^t} f_{\text{dis}}\left(\Theta, D_{\text{train}}, t\right) = \frac{1}{\left|D_{\text{train}}\right|}\sum_{n=1}^{\left|D_{\text{train}}\right|}\frac{\partial F\left(y_n \mid \mathbf{v}_n;\tilde{\mathbf{o}}_t\right)}{\partial\theta^t} - \sum_{y'} p\left(y' \mid \mathbf{v}_n;\tilde{\mathbf{o}}_t\right)\frac{\partial F\left(y' \mid \mathbf{v}_n;\tilde{\mathbf{o}}_t\right)}{\partial\theta^t}, \quad (30)$$

where,

$$F\left(y \mid \mathbf{v};\tilde{\mathbf{o}}_t\right) = -\ln\sum_{z=1}^{\infty}\left(e^{-G\left(y,z\mid\mathbf{v};\tilde{\mathbf{o}}_t\right)}\right), \quad (31)$$

and

$$p\left(y \mid \mathbf{v};\tilde{\mathbf{o}}_t\right) = e^{-F\left(y\mid\mathbf{v};\tilde{\mathbf{o}}_t\right)} \Big/ \sum_{y'} e^{-F\left(y'\mid\mathbf{v};\tilde{\mathbf{o}}_t\right)}. \quad (32)$$

As there are $l_t$ hidden units with non-zero weights at step *t,* the complexity of computing (32) is $\mathrm{O}\left(l_t^2 D + l_t^2 C\right)$.

The gradients $\partial F\left(y_t \mid \mathbf{v}_t;\tilde{\mathbf{o}}_t\right)/\partial\theta^t$ can be exactly computed, which are shown in the **Appendix**. However, this involves computing gradients for infinite many parameters. To avoid this issue, we only compute the gradients for parameters of first $l_t$ hidden units, and leave all the remaining parameters to be 0. This operation is equivalent to using (30) to compute the gradients for non-zero parameters and (29) for the remaining parameters.

The maximum number of activated hidden units $l_t$ changes gradually during training. Practically, if the Gibbs sampling chain ever samples a value of $z$ larger than $l_t$, we clamp it to $l_t + 1$. This avoids filling large memory for a large (but rare) value of $z$.

### 3.3 Evaluation of the models

After the training is done (with $T$ steps), we can simply treat $l_T$ as the number of hidden units been trained. However this results in many “redundant” hidden units, as many of weight vectors is fairly small.

As $P\left(z \mid \mathbf{v}_n\right)$ reflects how strong the filters respond to an input. We can use this information to estimate the size of efficiently trained hidden layer size as training progresses, which is given below

$$N_h \approx \frac{1}{M_b}\sum_{batch=1}^{N_b}\max\left\{\arg\max_z P\left(z \mid \mathbf{v}_n\right), \mathbf{v}_n \in D_{batch}\right\}, \quad (33)$$

where, $N_b$ is the number of mini-batches of the training set. $D_{batch}$ is the current mini-batch. In (33) we take an average between mini-batches of the maximal *z* giving the highest $P(z \mid \mathbf{v}_n)$ in each mini-batch.

Principally, the likelihood of a new data point $\mathbf{v}'$, $p(\mathbf{v}')$ for iRBM or $y$ conditioned on $\mathbf{v}'$, $p(y \mid \mathbf{v}')$ for Dis-iRBM are computed as follows:

$$p(\mathbf{v}') = \mathbb{E}_{p(\mathbf{o}_T)}\left[p(\mathbf{v}' \mid \mathbf{o}_T)\right], \quad (34)$$

$$p(y \mid \mathbf{v}') = \mathbb{E}_{p(\mathbf{o}_T)}\left[p(y \mid \mathbf{v}'; \mathbf{o}_T)\right]. \quad (35)$$

However, computing (34) and (35) are much more expensive than simply computing $p(\mathbf{v}' \mid \mathbf{o}_T)$ and $p(y \mid \mathbf{v}'; \mathbf{o}_T)$, as there are $M_T!$ different permutations. A way to deal with it is using samples $\tilde{\mathbf{o}}_T$ : $p(\mathbf{v}') \approx \frac{1}{N}\sum_{n=1}^{N} p(\mathbf{v}' \mid \tilde{\mathbf{o}}_T^n)$, $p(y \mid \mathbf{v}') \approx \frac{1}{N}\sum_{n=1}^{N} p(y \mid \mathbf{v}'; \tilde{\mathbf{o}}_T^n)$.

But if $p(z \le M_T \mid \mathbf{v}, \tilde{\mathbf{o}}) \to 0$ for an arbitrary order of the first $M_T$ hidden units is satisfied, the likelihood $p(\mathbf{v})$ is invariant to the order of the first $M_T$ hidden units. Any order gives the same result, then $p(\mathbf{v}) = p(\mathbf{v}' \mid \tilde{\mathbf{o}}_T(1:M_T), \tilde{\mathbf{o}}_0(M_T+1, N_h))$, where $\tilde{\mathbf{o}}_T(1:M_T)$ is an arbitrary order of the first $M_T$ hidden units, $\tilde{\mathbf{o}}_0(M_T+1, N_h)$ is the original order of the remaining hidden units. This conclusion is formally represented in the following form:

**Proposition 3.1** For any permutation of the first $M_T$ hidden units $\tilde{\mathbf{o}}_T(1:M_T)$, if $p(z \le M_T \mid \mathbf{v}, \tilde{\mathbf{o}}_T(1:M_T)) \to 0$, $\forall \mathbf{v} \in \{0,1\}^D$ is satisfied, then

$$p(\mathbf{v}) = p\left(\mathbf{v} \mid \tilde{\mathbf{o}}_T^1(1:M_T), \tilde{\mathbf{o}}_0(M_T+1:N_h)\right) = p\left(\mathbf{v} \mid \tilde{\mathbf{o}}_T^2(1:M_T), \tilde{\mathbf{o}}_0(M_T+1:N_h)\right),$$

where $\tilde{\mathbf{o}}_T^1(1:M_T)$ and $\tilde{\mathbf{o}}_T^2(1:M_T)$ are two arbitrary permutations of the first $M_T$ hidden units, and $\tilde{\mathbf{o}}_0(M_T+1:N_h)$ is the original permutation of the remaining hidden units. The proof is given in the **Appendix**.

Experimental results have shown that RP training can successfully make the condition in Proposition 3.1 approximately satisfied, any ordering gives a nearly identical result, thus a small $N$ (e.g. $N$=5) is enough to give a good estimate of (34) and (35).

## 4 Experiments and discussions

In this section, we evaluate our training strategy empirically according to the convergence speed and the final generalization performance. The datasets used for evaluation are binarized MNIST [22] and CalTech101 Silhouettes [23].The MNIST dataset is composed of 70,000 images of size $28 \times 28$ pixels representing handwritten digits (0-9), among which 60,000 images are used for training, and 10,000 images for testing. Each pixel of the image has been stochastically binarized according to its intensity as in [22]. The CalTech101 Silhouettes dataset is composed of 8,671 images of size 28 × 28 binary pixels, representing object silhouettes of 101 classes. The dataset is divided into three parts: 4,100 examples for training, 2,264 for validation and 2,307 for testing.We reshape each image of both datasets into a 784-dimensional vector by concatenating the adjacent rows one by one.

We have designed several experiments for different purposes. In Subsect. 4.1, the principle of choosing a proper regrouping rate $M_t$ was experimentally investigated. In Subsect. 4.2, we evaluated the generalization performance of the iRBM trained with RP according to its log-likelihood on the test sets of binarized MNIST and CalTech101 Silhouettes. In Subsect. 4.3, we evaluated the generalization performance of Dis-iRBMs trained with RP on classification tasks. For all the experiments, the mini-batch size is 100 and (P)CD is used to compute the gradients. Max-norm regularization [16] was also used to suppress very large weights, the bounds for each $\mathbf{W}_{i\cdot}$ and $\mathbf{U}_{i\cdot}$ were 10 and 5 respectively. Côté and Larochelle [14] claims that results of learning are robust to the value of the hidden unit penalty $\beta_i$. We have tried several different $\beta_i$ and find that smaller $\beta_i$ enables the model to grow to proper size faster at the beginning of learning. However, due to the ordering effect, it takes a long time for the hidden units to learn filters diverse from each other. RP training also prefers a small $\beta_i$ to allow as many hidden units to be mixed as possible. But a too small $\beta_i$ (coupled with a large $M_t$ ) is more likely to cause the model grow explosively. Based on the above arguments, and for convenience of comparing, we used the same $\beta_i = 1.01 \times \ln 2$ for all the models in this paper, which is identical to that in [14]. We also used L1 regularization and L2 regularization to regularize the models. The code to reproduce the results of this paper is available on GitHub[1].

We would like to mention that there exist a number of sophisticated techniques that improve performance of classic RBMs on sampling strategies [24， 25], model architectures [26，27], etc. However, the aim of this paper is to propose a alternative training strategy for faster convergence and better generalization of the original iRBMs in general. Combining these techniques can benefit for both training strategies, here we focus on comparing on basic settings of parameters.

[1]https://github.com/Boltzxuann/RP-iRBM

### 4.1 The principle of choosing the regrouping rate $M_t$

As mentioned in Sect. 3, choosing a proper $M_t$ is essential for RP training. In this subsection, we experimentally investigated the influence of $M_t$ on the growing of the model, and a principle of choosing it is proposed based on the experimental results.

In order to have a preliminary understanding on how different regrouping rates $M_t$ influence the growing of the model, we have tried several different settings of $M_t$ from 0 to $0.9l_t$. The model was trained for 10 epochs on binarized MNIST and 60 epochs on CalTech101 Silhouettes for each setting, and the training was repeated for 5 times. The mean results of the 5-time trials are illustrated in Fig. 4. The results of $M_t = 0$ illustrate the growing of iRBMs without RP training, which are the baselines for comparison. As shown in Fig. 4, the growing of the model is not remarkably influenced as long as $M_t$ is smaller than $0.8l_t$. The growing is even more stable when using RP for CalTech101 Silhouettes.

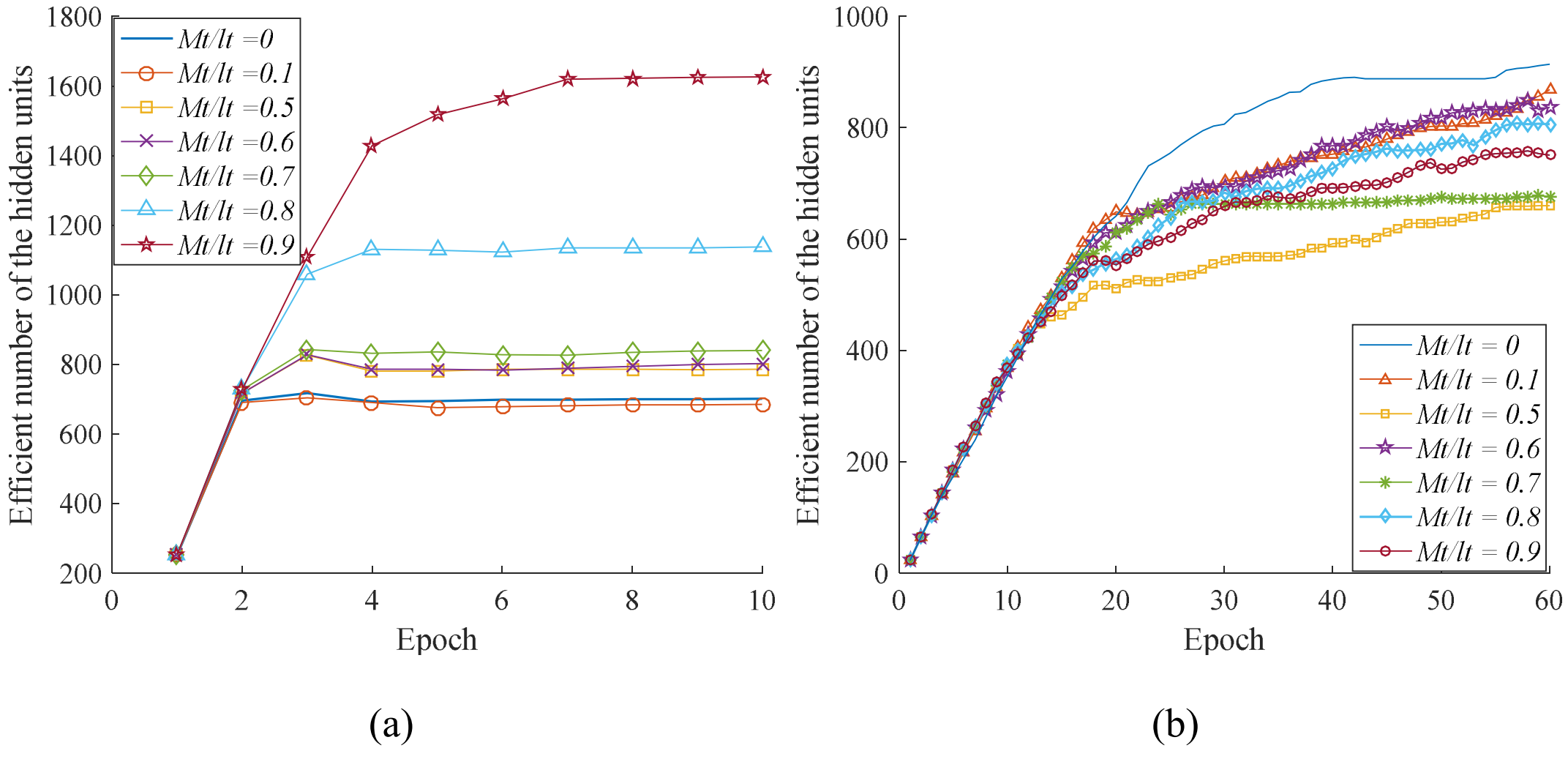

Fig.4. Growing of iRBMs when using different regrouping rates $M_t$, (a) binarized MNIST; (b) CalTech101 Silhouettes.

Fig. 5 illustrates the results of $p(z \mid \mathbf{v}_n; \tilde{\mathbf{o}}_T)$ for all training examples after 50 epochs of training on binarized MNIST and 300 epochs of training on CalTech101 Silhouettes. An interesting fact of RP trained iRBMs is that, RP training pushes all the $z$ that maximize $p(z \mid \mathbf{v}_n; \tilde{\mathbf{o}}_T)$ above $M_T$. Similar numbers of hidden units are activated even though the inputs are quite different. The number of activated hidden units ranges from 500~520 on binarized MNIST and 700~750 on CalTech101 Silhouettes.

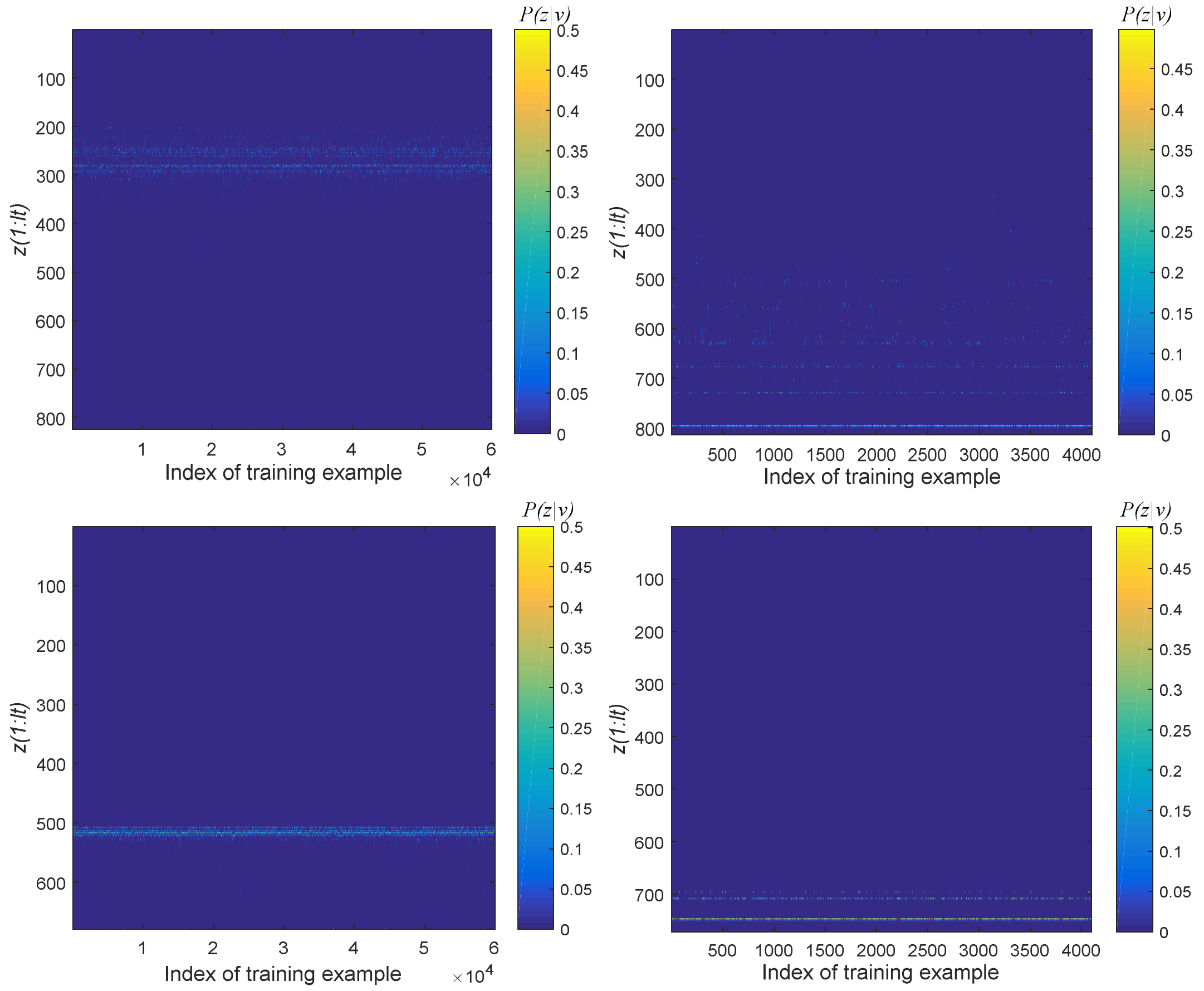

Fig. 5. A illustration of $p(z \mid \mathbf{v}_n)$ for all training examples after 50 epochs of training on binarized MNIST (left column) and 300 epochs of training on CalTech101 Silhouettes (right column). The top row are results without RP training, while the bottom row are those with RP training.

Proposition 3.1 states that if $p(z \le M_T \mid \mathbf{v}, \tilde{\mathbf{o}}_T) \to 0$ for all $\mathbf{v} \in \{0,1\}^D$ is fulfilled, $p(\mathbf{v})$ is invariant to the order of the first $M_T$. In that case, $p(z \le M_T \mid \mathbf{v}_n, \tilde{\mathbf{o}}_T) \to 0$ for all observed data $\mathbf{v}_n$. This can be easily checked by computing $p(z \le M_T \mid \mathbf{v}_n, \tilde{\mathbf{o}}_T)$ for all training examples on an arbitrary permutation $\tilde{\mathbf{o}}_T$. After 50 epochs of training on binarized MNIST and 300 epochs on CalTech101 Silhouettes, the average value of $\ln p(z \le M_T \mid \mathbf{v}_n, \tilde{\mathbf{o}}_T)$ on the two training sets are -28.06 and -37.46 respectively, which are fairly small. The condition in Proposition 3.1 is approximately satisfied. Plots of $p(z \le i \mid \mathbf{v}_n, \tilde{\mathbf{o}}_T)$ from $i = 1$ to $l_t$ for all training examples on the two datasets are shown in Fig. 6.

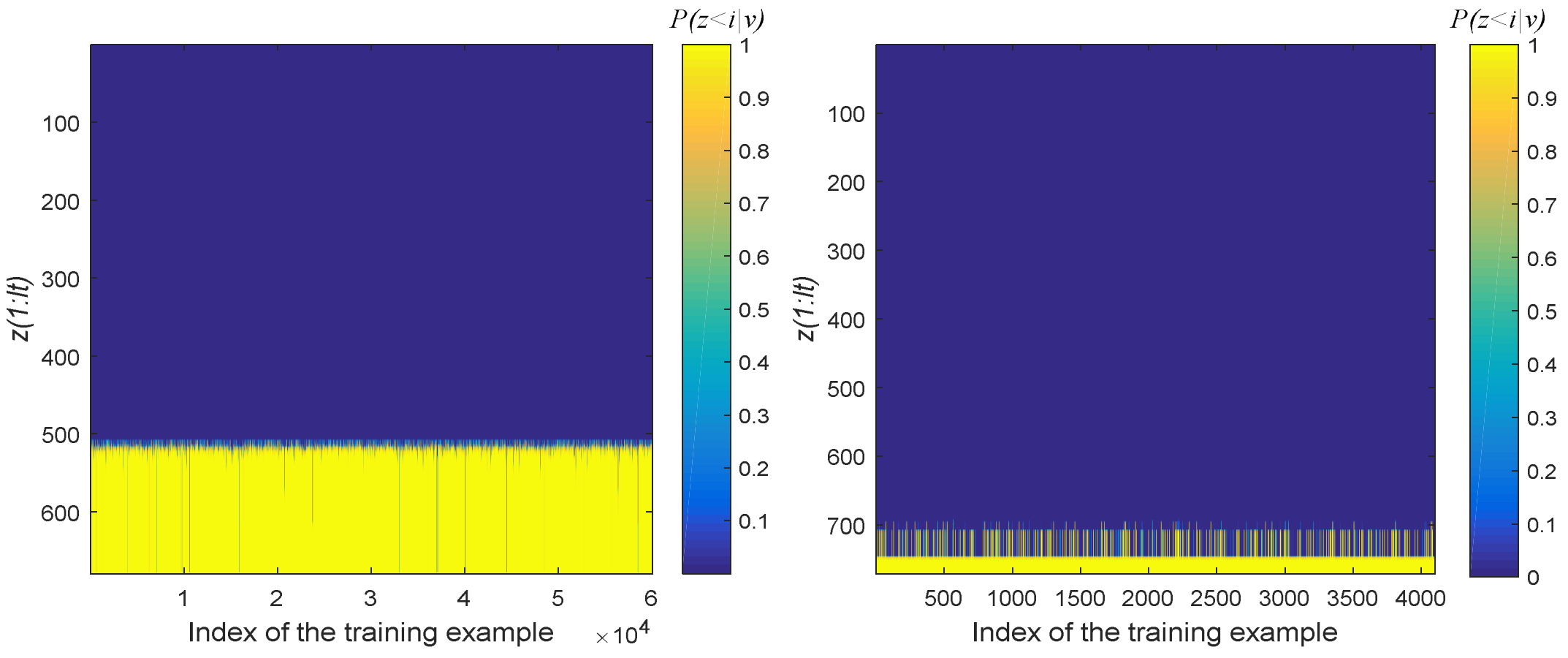

Fig. 6. A illustration of $p\left(z<i \mid \mathbf{v}_n\right)$ for all training examples after 50 epochs of training on binarized MNIST (left) and 300 epochs of training on CalTech101 Silhouettes (right). $p\left(z<M_t \mid \mathbf{v}_n\right) \rightarrow 0$ are approximately satisfied on the whole training sets.

As $p\left(z \mid \mathbf{v}_n\right)$ is influenced by the value of $M_t$, it can be used as "feedback" to guide the change of $M_t$ during training, which lead to a more adaptable $M_t$ proposed as follows:

$$M_t \approx \frac{1}{0.2 \times epoch} \sum_{0.8 \times epoch}^{epoch-1} M_z\left(epoch'\right) - 10 \quad , \tag{32}$$

where $M_z\left(epoch\right)$ is the average number of activated hidden units defined below:

$$M_z\left(epoch\right) = \frac{1}{\left|D_{\text{train}}\right|} \sum_{n=1}^{\left|D_{\text{train}}\right|} \arg\max_z p\left(z \mid \mathbf{v}_n ; \tilde{\mathbf{o}}_{t \in epoch}\right) \quad . \tag{33}$$

Based on the above analysis, a principle of choosing $M_t$ is proposed here:

At the beginning of training, $M_t = 0.7 l_t \sim 0.8 l_t$ to allow a greedy mixing of iRBMs. After that, a more adaptable $M_t$ defined by (32) is used to stabilize the growing.

### 4.2 Evaluating the generalization performance of RP trained iRBMs as density models

This subsection aims at evaluating the generalization performance of the iRBM trained with RP as a density model. Firstly, we investigated the property of RP training with different learning rates and regularizations. We have also trained an iRBM without RP training, as a comparison to our method. All the iRBMs were trained on binarized MNIST. Two different learning rates, a decaying learning rate ( $lr \propto 1/t$ ) and an adaptive learning rate method ADAGRAD [28], were used. And the global learning rate for ADAGRAD was 0.05, which gives the best results in [14]. The weights of the regularizations (L1 and L2) were set as $1\times10^{-4}$. Each model was trained up to 1000 epochs and we performed the Annealed Importance Sampling [22] to estimate the log likelihood of the test set every 50 epochs. The results are shown in Fig. 7. For ordinarily trained iRBMs, we only show the results using ADAGRAD, as the decaying learning rate leads to fairly poor convergence. After 1000 epochs of training, the effective activated number of hidden units of the ordinary trained iRBMs using CD are 805 (L2 regularization) and 566 (L1 regularization), which are larger than those of RP trained iRBMs (588 for L2 regularization and 501 with L1 regularization). And their convergence is much slower than the later. Due to the "ordering effect", many filters of the former are inadequately trained, which results in more "redundant" hidden units than the later. It is also noticeable that L1 regularization gives better results than L2 regularization. This may indicate that sparsity can lead to better generalization for iRBMs.

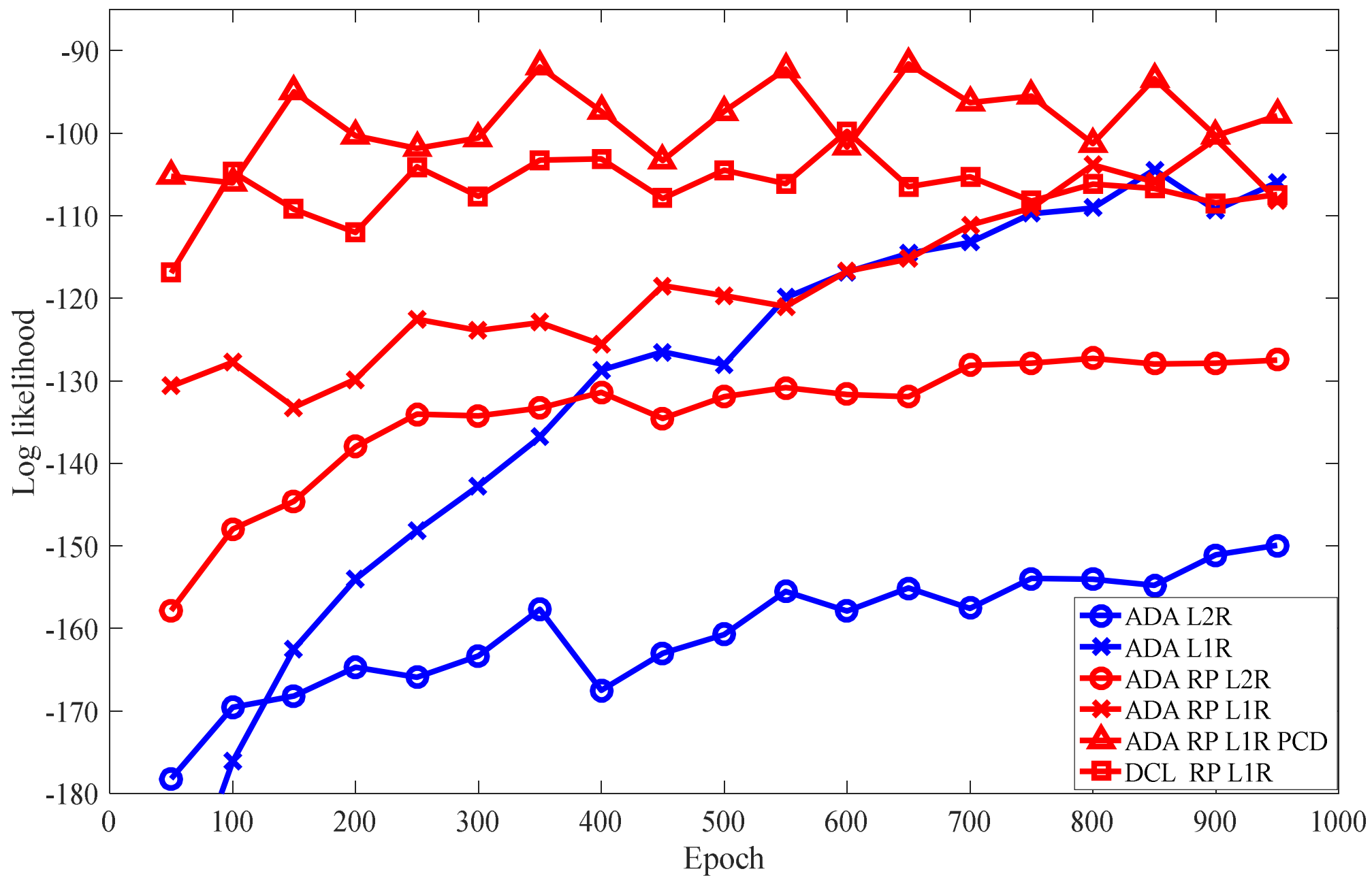

Fig. 7. Log-likelihood on test set of binarized MNIST for iRBMs with different training strategies. Where "L1R" and "L2R" stand for "L1 regularization" and "L2 regularization" respectively, "ADA" stands for "ADAGRAD", "DCL" stands for "decaying learning rate".

Fig. 8 illustrates the filters learned by the hidden units of the two iRBMs after 10 epochs of training, where the left-most 100 filters are illustrated. As shown in Fig. 8, the filters learned by hidden units with RP training are more diverse from each other than those learned without RP training. The former contains various kinds of local features such as strokes and specific character parts. While the ordering effect in the later is obvious, the left filters look like mixtures of different characters, and the right ones are just different shapes of the character "1".

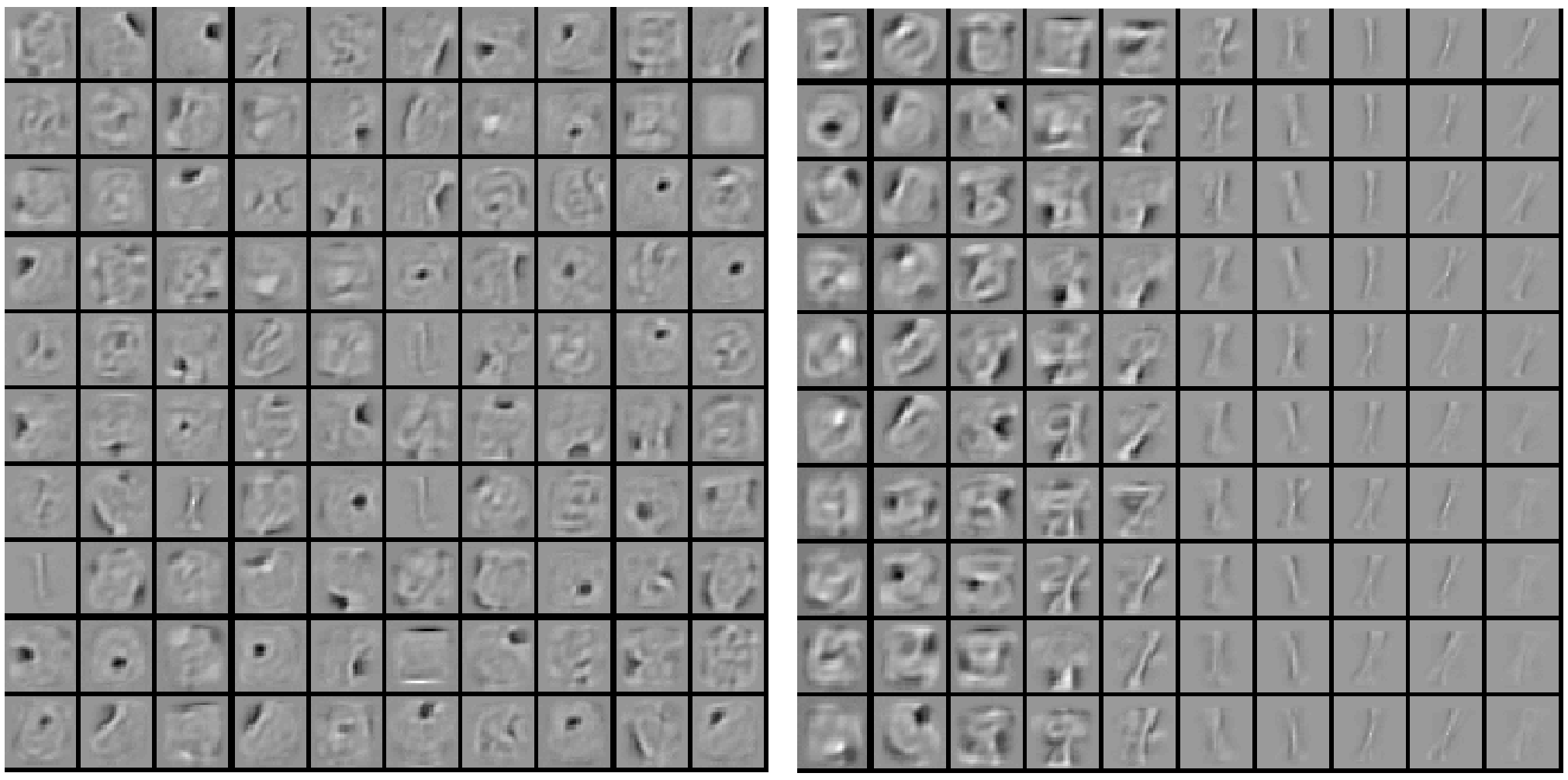

(a) (b)

Fig. 8. Comparison of learned filters between iRBMs with different learning strategies after 10 epochs of training, (a) with RP; (b) without RP. The left-most100 filters are shown starting from the top-left corner and incrementing across rows first.

Based on the above experimental results, we used PCD with 10 steps of Gibbs sampling and CD with 25 steps of Gibbs sampling to train the models on binarized MNIST and CalTech101 Silhouettes respectively. And L1 regularization was used all models. Each model was trained up to 1000 epochs. We have also performed a grid search on all the hyper parameters (global learning rate: 0.001~1, L1 regularization weight: 0~0.01).The best results of different models are illustrated in Table 1.

Table 1. The best results of average log-likelihood on test sets of binarized MNIST and CalTech101 Silhouettes of different models.

| Binarized MNIST | | | CalTech101 Silhouettes | | |
|---|---|---|---|---|---|
| Model | Size | Avg. LL | Model | Size | Avg. LL |
| RBM [22] | 500 | -86.34 | RBM [14] | 500 | -119.05 |
| iRBM [14] | 1208 | -85.65 | iRBM [14] | 915 | -121.47 |
| FW-iRBM [18] | 460 | ≈-85 | FW-iRBM [18] | 550 | ≈-127 |
| iRBM, RP | **674** | **-85.81** | iRBM, RP | **695** | **-114.09** |

The best model trained on binarized MNIST has 674 effective hidden units. The global learning rate is 0.05, and L1 regularization weight is $1\times10^{-4}$. Its average log-likelihood on the test set is -85.81, which is similar to-85.65 (3000 epochs of training) in [14]. The size of our model is also smaller than that in [14], which has 1208 hidden units with non-zero parameters. The best model trained on CalTech101 Silhouettes has 695 effective hidden units. The global learning rate is 0.02, and L1 regularization weight is $1\times10^{-3}$. Its average log-likelihood on the test set is -114.09, which is better than -121.47 in [14]. The results of FW-iRBM [18] are also listed in the table, the best results are achieved by RBMs with CD training initiated by FW-iRBMs. Samples from the best models are illustrated in Fig. 9.

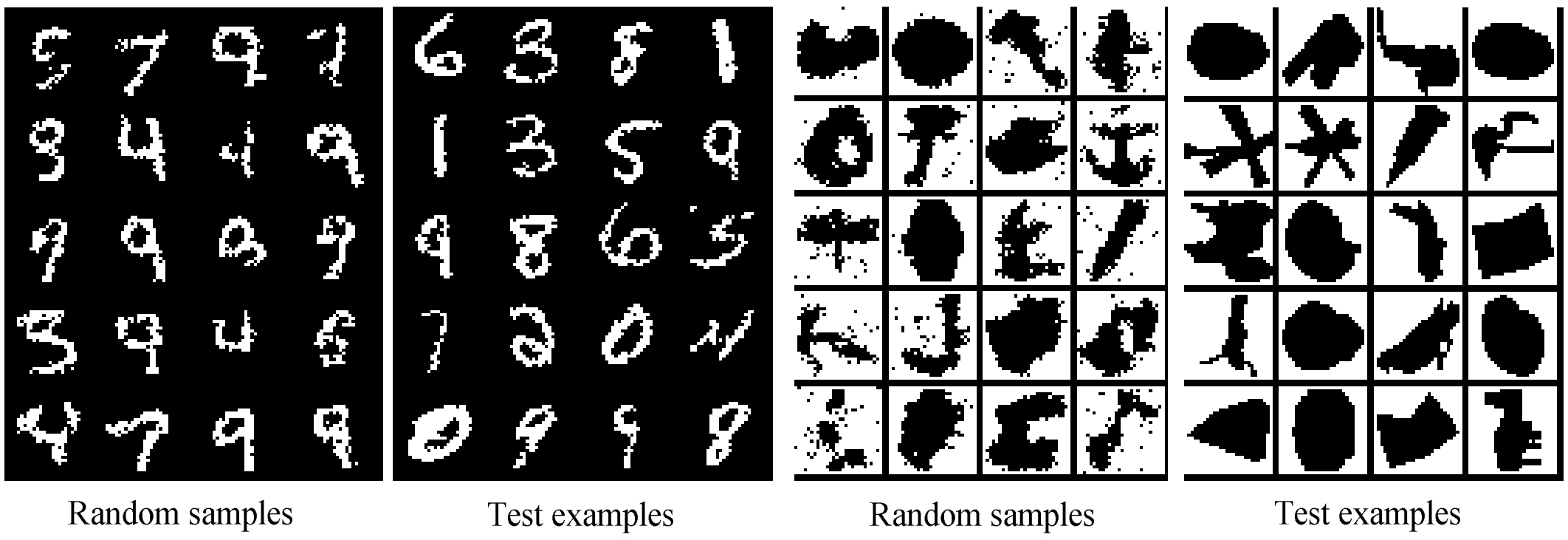

(a) (b)

Fig. 9. Random samples drawn from the best models by randomly initializing visible units and running10,000 Gibbs steps, examples from the test set are also illustrated, (a) binarized MNIST; (b) CalTech101 Silhouettes.

To show how RP training affects the dependency between $z$ and $\mathbf{v}$. We computed the histogram of $z_m = \arg\max_z p(z \mid \mathbf{v}_t) \approx \sum_{n=1}^{5} p(z \mid \mathbf{v}_t; \tilde{\mathbf{o}}_T^n)$ on the two test sets. The results are shown in Fig. 10, which reveals two facts: (a) All $z_m$ are larger than $M_t$; (b) All the inputs have similar numbers of activated hidden units, as all $z_m$ are close to each other. The number of example-specific filters has been greatly reduced.

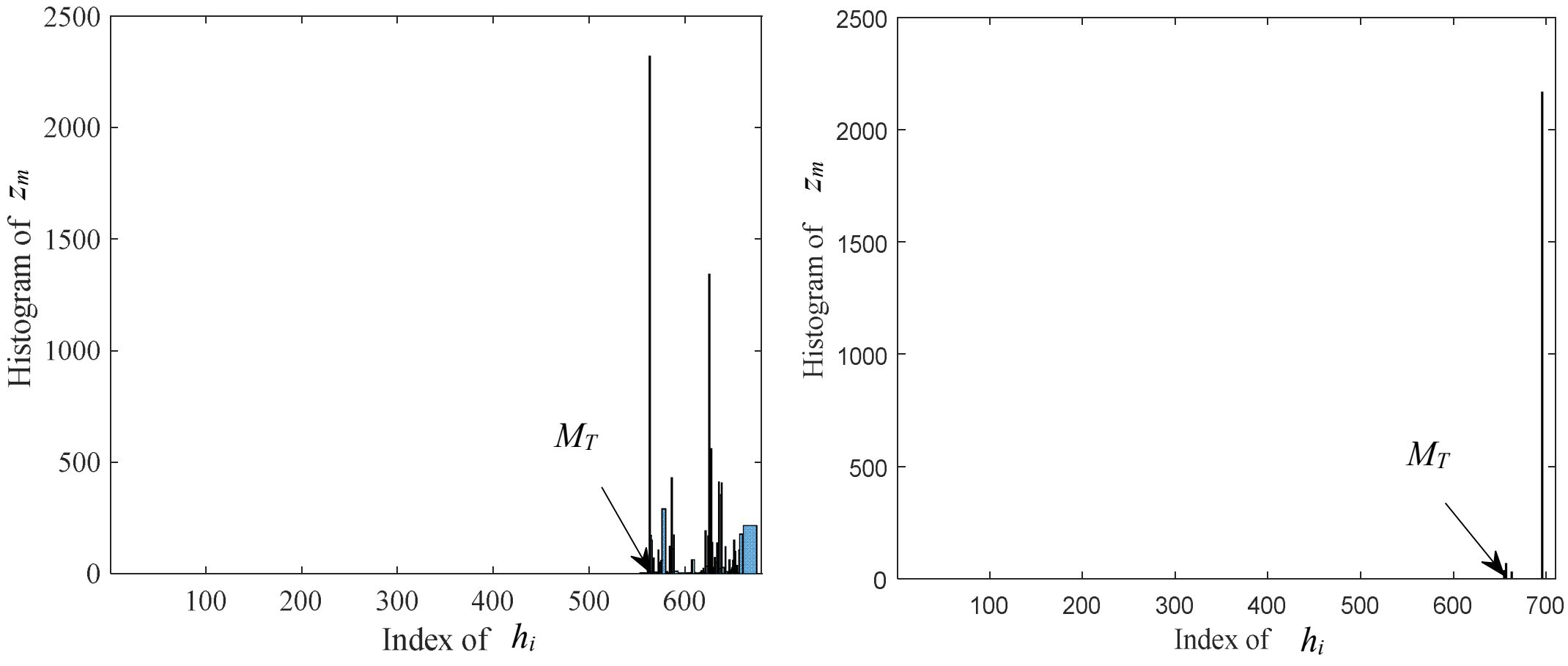

(a) (b)

Fig. 10. The histogram of $z_m$ on the two test sets, (a) binarized MNIST; (b) CalTech101 Silhouettes.

To further validate that the filters are independent from each other, we just use all the learned filters to compose a classic RBM, i.e. $z$ is clamped to $N_h$ , $p(\mathbf{v}) \approx p(\mathbf{v} \mid N_h)$ . The average log-likelihood on the two test sets are -88.13(binarized MNIST) and -115.90 (CalTech101 Silhouettes) for the converted RBMs.

### 4.3 Evaluating the generalization performance of RP trained Dis-iRBMs on classification tasks

In this subsection, we aim at evaluating the generalization performance of Dis-iRBMs trained with RP on binarized MNIST and CalTech101 Silhouettes. Firstly we compared the performance of RP with ordinary training to validate its boosting on learning speed and as guidance for selecting hyper-parameters. The discriminative training objective ( $\alpha = 0$ ) was optimized. Two different learning rates ( $lr = 0.1$ , 1) together with ADAGRAD (global learning rate is 0.1), were used. For each learning rate, we trained the model with or without our training strategy for 5 epochs. At the end of every epoch, miss-classification rate on the test set was computed. We repeated the training procedure for each parameter setting 5 times. The results are shown in Fig. 10. As shown in Fig. 11, RP accelerates learning for all the learning rates. ADAGRAD coupled with RP training achieves the best performance in this experiment. The best result with RP training at epoch 5 is 3.10%. High learning rate (e.g., $lr = 1$ ) encourages the model to quickly explore different regions of the weight space, but it is also more likely to cause oscillating unless proper learning rate decay is used. ADAGRAD gives smaller learning rates to parameters close to convergence, thus it is more stable than the fixed learning rate. Another fact can be observed from Fig. 11 is that, RP training makes the learning more stable, the variance of learning is much smaller. As the additional label information makes the filters prefer some classes instead of others, the "ordering effect" is more significant.

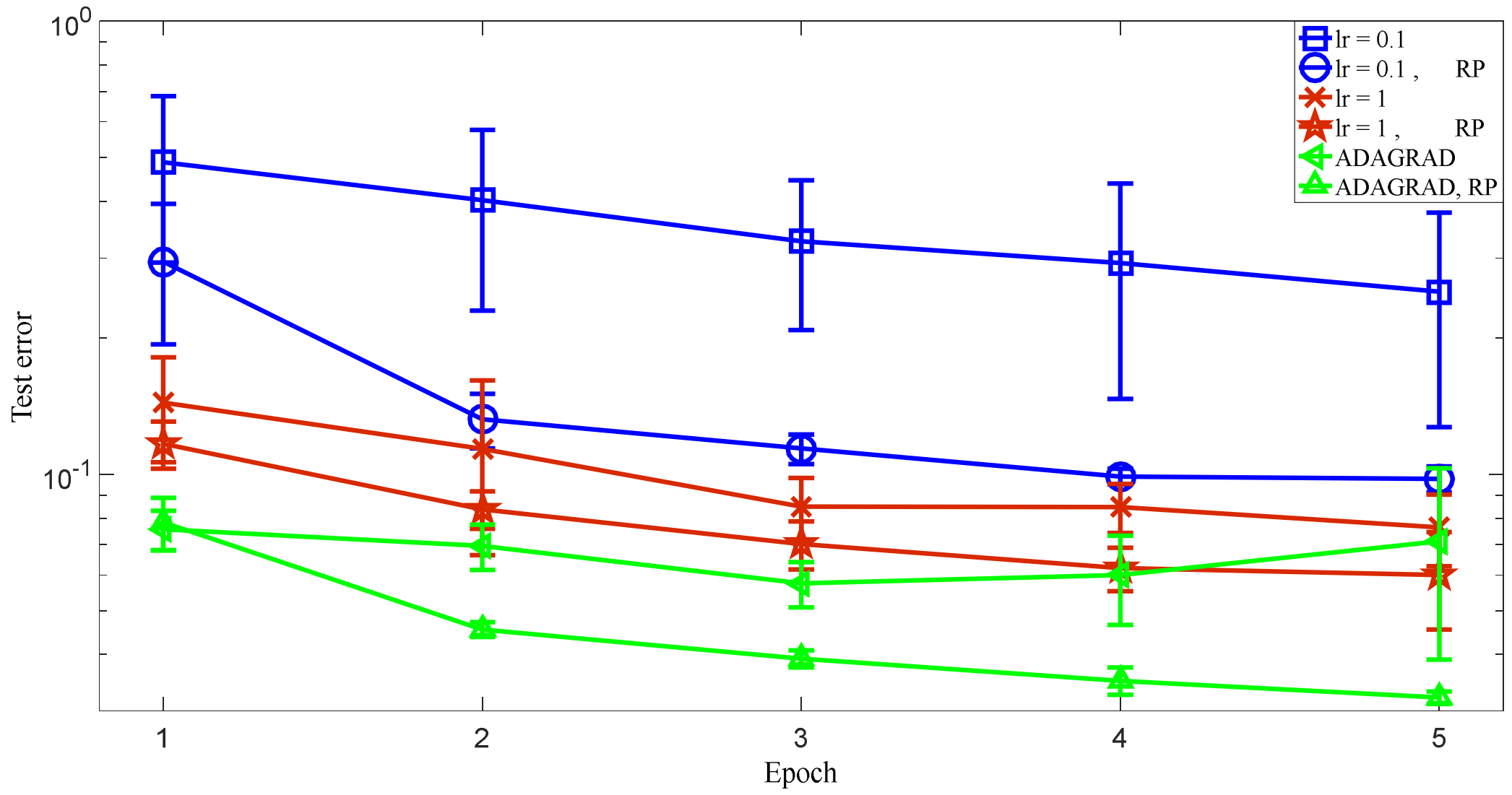

Fig. 11. Comparison of convergence speed between Dis-iRBMs trained with or without RP. The test errors are illustrated in log-scale.

According to the results of the above experiment, we used RP to train a Dis-iRBM on binarized MNIST and CalTech101 Silhouettes, and evaluated their generalization performance according to the test miss-classification errors. ADAGRAD were used in all experiments. The validation sets were used to search the hyper parameters on a log-scale (the generative proportion $\alpha$ : 0~1, global learning rate: 0.001~1, L1 regularization weight: 0~0.01). The best results for different models and method are illustrated in Table 2.

Table 2. The best results of classification error on the test set of MNIST and CalTech101 Silhouettes achieved by different models.

| Binarized MNIST | | | | CalTech101 Silhouettes | | | |
|---|---|---|---|---|---|---|---|
| Model | Size | Training objective | Test error | Model | Size | Training objective | Test error |
| ClassRBM [19] | 500 | Dis. | 1.81% | Dis-iRBM | 235 | Dis. | 37.10% |
| | 1500 | Hybrid ($\alpha = 0.01$) | 1.28% | | 373 | Hybrid ($\alpha = 0.01$) | 34.59% |
| Dis-iRBM | 382 | Dis. | 2.20% | FW-iRBM+SR [18] | 600 | -- | 34.5% |
| | 416 | Hybrid ($\alpha = 0.005$) | 1.71% | Dis-iRBM, RP | **165** | Dis. | **34.55%** |
| FW-iRBM+SR [18] | 600 | -- | 2.2% | | **742** | Hybrid ($\alpha = 0.01$) | **30.95%** |
| Dis-iRBM, RP | **489** | Dis. | **1.92%** | | | | |
| | **621** | Hybrid ($\alpha = 0.005$) | **1.41%** | | | | |

The best Dis-iRBM trained using hybrid training objective ( $\alpha = 0.005$ ) achieves a test error of 1.42% on MNIST, which is better than 1.71% achieved by normally trained Dis-iRBM. The global learning rate for it is 0.1, and the L1 regularization weight is $1\times10^{-4}$. Dis-iRBM performs slightly worse than the ClassRBM on MNIST , but the difference between the two best results is smaller than 0.2%, which is commonly regarded as statistical insignificant for MNIST. The best result on CalTech101 Silhouettes is also achieved by a Dis-iRBM trained using hybrid training objective ( $\alpha = 0.01$ ) and RP training. The test error is 30.95%, which is better than 34.59% achieved by normally trained Dis-iRBM. The global learning rate for the best model is 0.01, and the L1 regularization weight is $1\times10^{-3}$ . An interesting fact about the model size of iRBMs trained with different training objective is that size is smaller when merely the discriminative training objective is used. This makes sense as less features are often needed if the model only needs to discriminate objects from each other, instead of modeling all the examples well.

The FW-iRBM [18] have also been used for classification by taking the hidden units' activation vectors and using them as input for a softmax regression (SR). As FW-iRBM cannot perform discriminative training directly due to its training objective, learning the FW-iRBM and learning the SR are two separate procedures. i.e., after training the FW-iRBM with $T$ iterations, fix the parameters and using its hidden units' activation vectors to train a SR. The results of FW-iRBM are also listed in the table.

A "trick" to make the learning a bit faster is using the momentum, we use different momentum values for parameters of different hidden units according to the time they have been trained. When the hidden unit is added to training, the starting momentum value is 0.5, and then it gradually increases to 0.9.

## 5 Conclusion and future work

In this paper, we have proposed a novel training strategy (RP training) for the infinite RBMs, which aims at achieving better convergence and generalization. The core concept of the RP training is a dynamic training objective that allows a different model to be optimized at each gradient descent step. More specifically, an iRBM with an random grouping of hidden units is sampled before doing gradient descent. An implicit mixture of infinite many iRBMs with different permutations of hidden units is achieved with RP training. Experiments on binarized MNIST and CalTech101 Silhouettes have shown that, RP can train the hidden units more efficiently, thus results in smaller hidden layer size and better generalization performance. Compared with the FW-iRBM, the iRBM trains all the hidden units jointly not one unit greedily each update step, thus the former is more likely to reach sub-optimal solution. In the future, more datasets especially some real-valued datasets, will be used to give a further evaluation of the performance of our training strategy. Meanwhile, we are exploring a multi-layer extension of the iRBM, the idea of RP training can be also applied to this new architecture, combined with a greedy layer-wise pre-training.

## Appendix

### A.1

**Algorithm 1** Parameter update of the iRBM for generative training objective with RP training

**Notation:** $\mathbf{x} \sim p$ means $\mathbf{x}$ is sampled from $p$

$\mathbf{x} \leftarrow \mathbf{y}$ means $\mathbf{x}$ is set to $\mathbf{y}$

**Initiation of parameters:** $l_0 \leftarrow 1$; $\forall \theta^0 \in \Theta_t$, $\theta^0 \leftarrow 0$

**for** $t, 2, \cdots T$ **do**

**Input:** training example $(\mathbf{v}_n, y_n)$, learning rate $lr(t)$, re-permutating length $M_t$

#Sampling a permutation of length $M_t$

$\tilde{\mathbf{o}}_t = [\tilde{o}_1, \tilde{o}_2, \cdots, \tilde{o}_{Mt}] \sim p_t(\mathbf{o}_t)$

#Re-permutating the hidden units and the corresponding parameters

$[h_1^t, h_2^t, \cdots, h_{Mt}^t] \leftarrow [h_{\tilde{o}_1}, h_{\tilde{o}_2}, \cdots, h_{\tilde{o}_{Mt}}]$

$[\theta_1^{t-1}, \cdots, \theta_{Mt}^{t-1}] \leftarrow [\theta_{\tilde{o}_1}^{t-1}, \cdots, \theta_{\tilde{o}_{Mt}}^{t-1}]$

#Positive phase of CD

$z^{\text{pos}} \sim p(z \mid \mathbf{v}_n; \tilde{\mathbf{o}}_t)$, $\hat{\mathbf{h}}^{\text{pos}} \leftarrow p(\mathbf{h} \mid \mathbf{v}_n, z^{\text{pos}}; \tilde{\mathbf{o}}_t)$

#Negative phase of CD

$\mathbf{h}^{\text{neg}} \sim p(\mathbf{h} \mid \mathbf{v}_n, z^{\text{pos}}; \tilde{\mathbf{o}}_t)$, $\mathbf{v}^{\text{neg}} \sim p(\mathbf{v} \mid \mathbf{h}^{\text{neg}}, z^{\text{pos}}; \tilde{\mathbf{o}}_t)$, $z^{\text{neg}} \sim p(z \mid \mathbf{v}^{\text{neg}}; \tilde{\mathbf{o}}_t)$

$\hat{\mathbf{h}}^{\text{neg}} \leftarrow p(\mathbf{h} \mid \mathbf{v}^{\text{neg}}, z^{\text{neg}}; \tilde{\mathbf{o}}_t)$

#Update

**for** $\theta^t \in \Theta_t$ **do**

$$\theta^t \leftarrow \theta^{t-1} - lr(t)\left(\frac{\partial}{\partial\theta} F(\mathbf{v}_n, z^{\text{pos}} | \tilde{\mathbf{o}}_t) - \frac{\partial}{\partial\theta} F(\mathbf{v}^{\text{neg}}, z^{\text{neg}} | \tilde{\mathbf{o}}_t)\right)$$

**end for**

#Adding a hidden unit

**if** $z^{\text{pos}} > l_{t-1}$ and $z^{\text{neg}} > l_{t-1}$ **then**

$l_t \leftarrow l_{t-1} + 1$

$\theta_{lt}^t \leftarrow 0$

**else then**

$l_t \leftarrow l_{t-1}$

**end if**

**end for**

## A.2

Derivation of gradients (30) w.r.t. $\theta^t$ is presented as follows. Without loss of generality, we only show the results related to the binary Dis-iRBM.

Recall the equation (30):

$$\frac{\partial}{\partial\theta^t} f_{\text{dis}}(\Theta, D_{\text{train}}, t) = \frac{1}{|D_{\text{train}}|} \sum_{n=1}^{|D_{\text{train}}|} \frac{\partial F(y_n \mid \mathbf{v}_n; \tilde{\mathbf{o}}_t)}{\partial\theta^t} - \sum_{y'} p(y' \mid \mathbf{v}_n; \tilde{\mathbf{o}}_t) \frac{\partial F(y' \mid \mathbf{v}_n; \tilde{\mathbf{o}}_t)}{\partial\theta^t},$$

where,

$$
\begin{aligned}
\frac{\partial F\left(y \mid \mathbf{v}_n;\tilde{\mathbf{o}}_t\right)}{\partial\theta} &= \frac{\partial}{\partial\theta} - \ln\sum_{z=1}^{\infty}\left(e^{-G\left(y,z\mid\mathbf{v}_n;\tilde{\mathbf{o}}_t\right)}\right) \\
&= -\frac{1}{\sum_{z=1}^{\infty}\left(e^{-G\left(y,z\mid\mathbf{v}_n;\tilde{\mathbf{o}}_t\right)}\right)}\frac{\partial}{\partial\theta}\sum_{z=1}^{\infty}\left(e^{-G\left(y,z\mid\mathbf{v}_n;\tilde{\mathbf{o}}_t\right)}\right) \\
&= \sum_{z=1}^{\infty}\frac{e^{-G\left(y,z\mid\mathbf{v}_n;\tilde{\mathbf{o}}_t\right)}}{\sum_{z=1}^{\infty}\left(e^{-G\left(y,z\mid\mathbf{v}_n;\tilde{\mathbf{o}}_t\right)}\right)}\frac{\partial}{\partial\theta}G\left(y,z\mid\mathbf{v}_n;\tilde{\mathbf{o}}_t\right) \\
&= \sum_{z=1}^{\infty}P\left(z\mid\mathbf{v}_n,y;\tilde{\mathbf{o}}_t\right)\frac{\partial}{\partial\theta}G\left(y,z\mid\mathbf{v}_n;\tilde{\mathbf{o}}_t\right)
\end{aligned}
\quad . \qquad (38)
$$

In order to compute (38), we need to compute $\frac{\partial}{\partial\theta}G\left(y,z\mid\mathbf{v}_n;\tilde{\mathbf{o}}_t\right)$ first, which are shown as follows:

$$
\begin{aligned}
\frac{\partial}{\partial\mathbf{W}_{i\cdot}^{t}}G\left(y,z\mid\mathbf{v};\tilde{\mathbf{o}}_t\right) &= -\frac{\partial}{\partial\mathbf{W}_{i\cdot}}\sum_{i'=1}^{z}\ln\left(1+\exp\left(\mathbf{W}_{i'\cdot}^{t}\mathbf{v}+\mathbf{U}_{i'\cdot}^{t}\mathbf{e}_y+c_{i'}^{t}\right)\right)-\beta_{i'} \\
&= -\frac{\partial}{\partial\mathbf{W}_{i\cdot}^{t}}\sum_{i'=1}^{z}\ln\left(1+\exp\left(\mathbf{W}_{i'\cdot}^{t}\mathbf{v}+\mathbf{U}_{i'\cdot}^{t}\mathbf{e}_y+c_{i'}^{t}\right)\right)-\beta_{i'} \\
&= -H\left(z-i\right)\frac{\partial}{\partial\mathbf{W}_{i\cdot}^{t}}\ln\left(1+\exp\left(\mathbf{W}_{i\cdot}^{t}\mathbf{v}+\mathbf{U}_{i\cdot}^{t}\mathbf{e}_y+c_{i}^{t}\right)\right) \\
&= -H\left(z-i\right)\frac{\exp\left(\mathbf{W}_{i\cdot}^{t}\mathbf{v}+\mathbf{U}_{i\cdot}^{t}\mathbf{e}_y+c_{i}\right)}{1+\exp\left(\mathbf{W}_{i\cdot}^{t}\mathbf{v}+\mathbf{U}_{i\cdot}^{t}\mathbf{e}_y+c_{i}\right)}\mathbf{v} \\
&= -H\left(z-i\right)\mathrm{s}\left(\mathbf{W}_{i\cdot}^{t}\mathbf{v}+\mathbf{U}_{i\cdot}^{t}\mathbf{e}_y+c_{i}\right)\mathbf{v}
\end{aligned}
\quad . \qquad (39)
$$

Similarly,

$$
\frac{\partial G\left(y,z\mid\mathbf{v};\tilde{\mathbf{o}}_t\right)}{\partial\mathbf{U}_{i\cdot}^{t}} = -H\left(z-i\right)\mathrm{s}\left(\mathbf{W}_{i\cdot}^{t}\mathbf{v}+\mathbf{U}_{i\cdot}^{t}\mathbf{e}_y+c_{i}^{t}\right)\mathbf{e}_y \quad , \qquad (40)
$$

$$
\frac{\partial}{\partial\mathbf{d}^{t}}G\left(y,z\mid\mathbf{v};\tilde{\mathbf{o}}_t\right) = -H\left(z-i\right)\mathbf{e}_y \quad , \qquad (41)
$$

$$
\frac{\partial}{\partial c_{i}^{t}}G\left(y,z\mid\mathbf{v};\tilde{\mathbf{o}}_t\right) = -H\left(z-i\right)\mathrm{s}\left(\mathbf{W}_{i\cdot}^{t}\mathbf{v}+\mathbf{U}_{i\cdot}^{t}\mathbf{e}_y+c_{i}^{t}\right) \quad , \qquad (42)
$$

where,

$$
H\left(z-i\right) = \begin{cases} 0, & z<i \\ 1, & z\geq i \end{cases} .
$$

Substitute (39) ~ (42) into (38), we get

$$
\begin{aligned}
\frac{\partial F\left(y\mid\mathbf{v};\tilde{\mathbf{o}}_t\right)}{\partial\mathbf{W}_{i\cdot}^{t}} &= -\sum_{z=1}^{\infty}P\left(z\mid\mathbf{v},y\right)H\left(z-i\right)\mathrm{s}\left(\mathbf{W}_{i\cdot}^{t}\mathbf{v}+\mathbf{U}_{i\cdot}^{t}\mathbf{e}_y+c_{i}^{t}\right)\mathbf{v} \\
&= P\left(z\geq i\mid\mathbf{v},y\right)\mathrm{s}\left(\mathbf{W}_{i\cdot}^{t}\mathbf{v}+\mathbf{U}_{i\cdot}^{t}\mathbf{e}_y+c_{i}^{t}\right)\mathbf{v}
\end{aligned}
\quad , \qquad (43)
$$

$$\frac{\partial F\left(y \mid \mathbf{v};\tilde{\mathbf{o}}_t\right)}{\partial \mathbf{U}_{i\bullet}^t} = -\sum_{z=1}^{\infty} P\left(z \mid \mathbf{v}, y\right) H\left(z-i\right) \mathrm{s}\left(\mathbf{W}_{i\bullet}^t \mathbf{v} + \mathbf{U}_{i\bullet}^t \mathbf{e}_y + c_i^t\right)\mathbf{e}_y$$
$$= P\left(z \ge i \mid \mathbf{v}, y\right) \mathrm{s}\left(\mathbf{W}_{i\bullet}^t \mathbf{v} + \mathbf{U}_{i\bullet}^t \mathbf{e}_y + c_i^t\right)\mathbf{e}_y \quad , \qquad (44)$$

$$\frac{\partial F\left(y \mid \mathbf{v};\tilde{\mathbf{o}}_t\right)}{\partial \mathbf{d}^t} = -\sum_{z=1}^{\infty} P\left(z \mid \mathbf{v}, y; \tilde{\mathbf{o}}_t\right)\mathbf{e}_y = -\mathbf{e}_y \quad , \qquad (45)$$

$$\frac{\partial F\left(y \mid \mathbf{v};\tilde{\mathbf{o}}_t\right)}{\partial c_i^t} = -\sum_{z=1}^{\infty} P\left(z \mid \mathbf{v}, y;\tilde{\mathbf{o}}_t\right) H\left(z-i\right) \mathrm{s}\left(\mathbf{W}_{i\bullet}^t \mathbf{v} + \mathbf{U}_{i\bullet}^t \mathbf{e}_y + c_i^t\right)$$
$$= P\left(z \ge i \mid \mathbf{v}, y;\tilde{\mathbf{o}}_t\right) \mathrm{s}\left(\mathbf{W}_{i\bullet}^t \mathbf{v} + \mathbf{U}_{i\bullet}^t \mathbf{e}_y + c_i^t\right) \quad , \qquad (46)$$

where, $P\left(z \ge i \mid \mathbf{v}, y;\tilde{\mathbf{o}}_t\right) = 1 - \sum_{z'=1}^{i} P\left(z' \mid \mathbf{v}, y;\tilde{\mathbf{o}}_t\right)$.

## A.3

The proof of proposition 3.1 is given as follows:

Suppose the permutation of the first $M_T$ hidden units is $\tilde{\mathbf{o}}_T^1\left(1:M_T\right)$. The likelihood of $\mathbf{v}$ conditioned on this permutation is given below:

From the condition given in Proposition 3.1 we can derive that,

$$p\left(z \le M_T \mid \mathbf{v}, \tilde{\mathbf{o}}_T^1\left(1:M_T\right)\right) \to 0, \quad \forall\ \mathbf{v} \in \{0,1\}^D$$

$$\Updownarrow$$

$$p\left(z = i \mid \mathbf{v}, \tilde{\mathbf{o}}_T^1\left(1:M_T\right)\right) \to 0, \text{ for } i = 1,2,\cdots,M_T$$

$$\Updownarrow$$

$$\frac{\exp\left(-F\left(\mathbf{v}, z \mid \tilde{\mathbf{o}}_T^1\left(1:M_T\right)\right)\right)}{\sum_{z'=1}^{M_T}\exp\left(-F\left(\mathbf{v}, z' \mid \tilde{\mathbf{o}}_T^1\left(1:M_T\right)\right)\right) + \sum_{z'=M_T+1}^{\infty}\exp\left(-F\left(\mathbf{v}, z' \mid \tilde{\mathbf{o}}_0\left(M_T+1:N_h\right)\right)\right)} \to 0, \text{ for }$$
$$i = 1,2,\cdots,M_T$$

$$\Updownarrow$$

$$\frac{\ln\left(1+\exp\left(\mathbf{W}_{i\bullet}\mathbf{v}+c_i\right)\right)}{\sum_{z'=M_T+1}^{\infty}\exp\left(-F\left(\mathbf{v}, z' \mid \tilde{\mathbf{o}}_0\left(M_T+1:N_h\right)\right)\right)} \to 0, \text{ for } i = 1,2,\cdots,M_T$$

$$\Updownarrow$$

$$\frac{\exp\left(-F\left(\mathbf{v},z\,|\,\tilde{\mathbf{o}}_T^2\left(1:M_T\right)\right)\right)}{\sum_{z'=M_T+1}^{\infty}\exp\left(-F\left(\mathbf{v},z'\,|\,\tilde{\mathbf{o}}_0\left(M_T+1:N_h\right)\right)\right)}\rightarrow 0, \tag{47}$$

for $i=1,2,\cdots,M_T$ and any permutation $\tilde{\mathbf{o}}_T^2\left(1:M_T\right)\neq\tilde{\mathbf{o}}_T^1\left(1:M_T\right)$ and $\forall\ \mathbf{v}\in\{0,1\}^D$

$$\Updownarrow$$

$$p\left(z=i\,|\,\mathbf{v},\tilde{\mathbf{o}}_T^2\left(1:M_T\right)\right)\rightarrow 0,\ \text{for}\ \ i=1,2,\cdots,M_T$$

The above conclusion indicates that if any permutation $\tilde{\mathbf{o}}_T\left(1:M_T\right)$ satisfies that $p\left(z=i\,|\,\mathbf{v},\tilde{\mathbf{o}}_T\left(1:M_T\right)\right)\rightarrow 0$, then all the other permutations $\tilde{\mathbf{o}}_T'\left(1:M_T\right)$ also satisfy $p\left(z=i\,|\,\mathbf{v},\tilde{\mathbf{o}}_T'\left(1:M_T\right)\right)\rightarrow 0$.

Now, consider the likelihood $p\left(\mathbf{v}\,|\,\tilde{\mathbf{o}}_T\left(1:M_T\right)\right)$ for an arbitrary permutation $\tilde{\mathbf{o}}_T\left(1:M_T\right)$:

$$\begin{aligned}
&p\left(\mathbf{v}\,|\,\tilde{\mathbf{o}}_T\left(1:M_T\right)\right)\\
&=\frac{\sum_{z=1}^{\infty}\exp\left(-F\left(\mathbf{v},z\,|\,\tilde{\mathbf{o}}_T\left(1:M_T\right)\right)\right)}{\sum_{z'=1}^{\infty}\sum_{\mathbf{v}'}\exp\left(-F\left(\mathbf{v}',z'\,|\,\tilde{\mathbf{o}}_T\left(1:M_T\right)\right)\right)}\\
&=\frac{\sum_{z=1}^{M_T}\exp\left(-F\left(\mathbf{v},z\,|\,\tilde{\mathbf{o}}_T\left(1:M_T\right)\right)\right)+\sum_{z=M_T+1}^{\infty}\exp\left(-F\left(\mathbf{v},z\,|\,\tilde{\mathbf{o}}_0\left(M_T+1:N_h\right)\right)\right)}{\sum_{z'=1}^{M_T}\sum_{\mathbf{v}'}\exp\left(-F\left(\mathbf{v}',z'\,|\,\tilde{\mathbf{o}}_T\left(1:M_T\right)\right)\right)+\sum_{z'=M_T+1}^{\infty}\sum_{\mathbf{v}'}\exp\left(-F\left(\mathbf{v}',z'\,|\,\tilde{\mathbf{o}}_0\left(M_T+1:N_h\right)\right)\right)}.
\end{aligned} \tag{48}$$

Take (47) into account, we have

$$p\left(\mathbf{v}\,|\,\tilde{\mathbf{o}}_T\left(1:M_T\right)\right)=\frac{\sum_{z=M_T+1}^{\infty}\exp\left(-F\left(\mathbf{v},z\,|\,\tilde{\mathbf{o}}_0\left(M_T+1:N_h\right)\right)\right)}{\sum_{z'=M_T+1}^{\infty}\sum_{\mathbf{v}'}\exp\left(-F\left(\mathbf{v}',z'\,|\,\tilde{\mathbf{o}}_0\left(M_T+1:N_h\right)\right)\right)}\ . \tag{49}$$

Equation (49) indicates that $p\left(\mathbf{v}\,|\,\tilde{\mathbf{o}}_T\left(1:M_T\right)\right)$ is invariant to the permutation $\tilde{\mathbf{o}}_T\left(1:M_T\right)$. The marginalized likelihood $p\left(\mathbf{v}\right)$ now becomes:

$$
\begin{aligned}
p(\mathbf{v}) &= \sum_{\tilde{\mathbf{o}}_T} p\left(\mathbf{v} \mid \tilde{\mathbf{o}}_T(1:M_T)\right) p\left(\tilde{\mathbf{o}}_T(1:M_T)\right) \\
&= \sum_{\tilde{\mathbf{o}}_T} \frac{\sum_{z=M_T+1}^{\infty} \exp\left(F\left(\mathbf{v}, z \mid \tilde{\mathbf{o}}_0(M_T+1:N_h)\right)\right)}{\sum_{z'=M_T+1}^{\infty} \sum_{\mathbf{v}'} \exp\left(F\left(\mathbf{v}', z' \mid \tilde{\mathbf{o}}_0(M_T+1:N_h)\right)\right)} p\left(\tilde{\mathbf{o}}_T(1:M_T)\right) \\
&= \frac{\sum_{z=M_T+1}^{\infty} \exp\left(F\left(\mathbf{v}, z \mid \tilde{\mathbf{o}}_0(M_T+1:N_h)\right)\right)}{\sum_{z'=M_T+1}^{\infty} \sum_{\mathbf{v}'} \exp\left(F\left(\mathbf{v}', z' \mid \tilde{\mathbf{o}}_0(M_T+1:N_h)\right)\right)} \sum_{\tilde{\mathbf{o}}_T} p\left(\tilde{\mathbf{o}}_T(1:M_T)\right) \\
&= \frac{\sum_{z=M_T+1}^{\infty} \exp\left(F\left(\mathbf{v}, z \mid \tilde{\mathbf{o}}_0(M_T+1:N_h)\right)\right)}{\sum_{z'=M_T+1}^{\infty} \sum_{\mathbf{v}'} \exp\left(F\left(\mathbf{v}', z' \mid \tilde{\mathbf{o}}_0(M_T+1:N_h)\right)\right)} \\
&= p\left(\mathbf{v} \mid \tilde{\mathbf{o}}_0(1:M_T)\right)
\end{aligned}
$$